\documentclass{article}

\usepackage[final]{corl_2026} 

\usepackage{multicol}

\usepackage{xcolor}
\definecolor{darkgreen}{RGB}{0,100,0}
\definecolor{rai}{HTML}{6C48FD}
\usepackage{makecell}

\usepackage[dvipsnames]{xcolor}
\usepackage{amssymb}
\usepackage{amsmath}
\usepackage{graphicx}
\usepackage{adjustbox}
\usepackage{float}
\usepackage{booktabs}
\usepackage{caption}
\usepackage{diagbox}
\usepackage{multirow}
\usepackage{booktabs}
\usepackage{makecell}
\usepackage{array} 
\usepackage{tcolorbox}
\usepackage{courier} 
\newcolumntype{P}[1]{>{\centering\arraybackslash}p{#1}}
\newcolumntype{C}[1]{>{\centering\arraybackslash}p{#1}}
\usepackage{wrapfig}
\usepackage{enumitem}

\newcount\Comments  
\newcommand{\zifan}[1]{{\ifnum\Comments=1\textcolor{blue}{[zifan: #1]}\fi}}

\title{\textsc{ExpertGen}: Scalable Sim-to-Real Expert Policy Learning from Imperfect Behavior Priors}

\author{
Zifan Xu$^{1,2}$, Ran Gong$^{1}$, Maria Vittoria Minniti$^{1}$, Kausik Sivakumar$^{1}$,\\
Ahmet Salih Gundogdu$^{1}$, Eric Rosen$^{1}$,  Riedana Yan$^{1}$, Tushar Kusnur$^{1}$, Zixing Wang$^{1}$, Di Deng$^{1}$, \\
Peter Stone$^{2,3}$, Xiaohan Zhang$^{\ast,1}$, Karl Schmeckpeper$^{\ast,1}$ \\ [4pt]
{\normalsize $^{1}$Robotics and AI Institute, $^{2}$University of Texas at Austin, $^{3}$Sony AI, $^{\ast}$Equal Advising} \\
{\normalsize \urlstyle{same} \textcolor{rai}{
\href{https://pages.rai-inst.com/expertgen/}{pages.rai-inst.com/expertgen}}}
}

\begin{document}

\maketitle

\begin{center}
    \vspace{-20pt}
    {\includegraphics[width=0.85\textwidth]{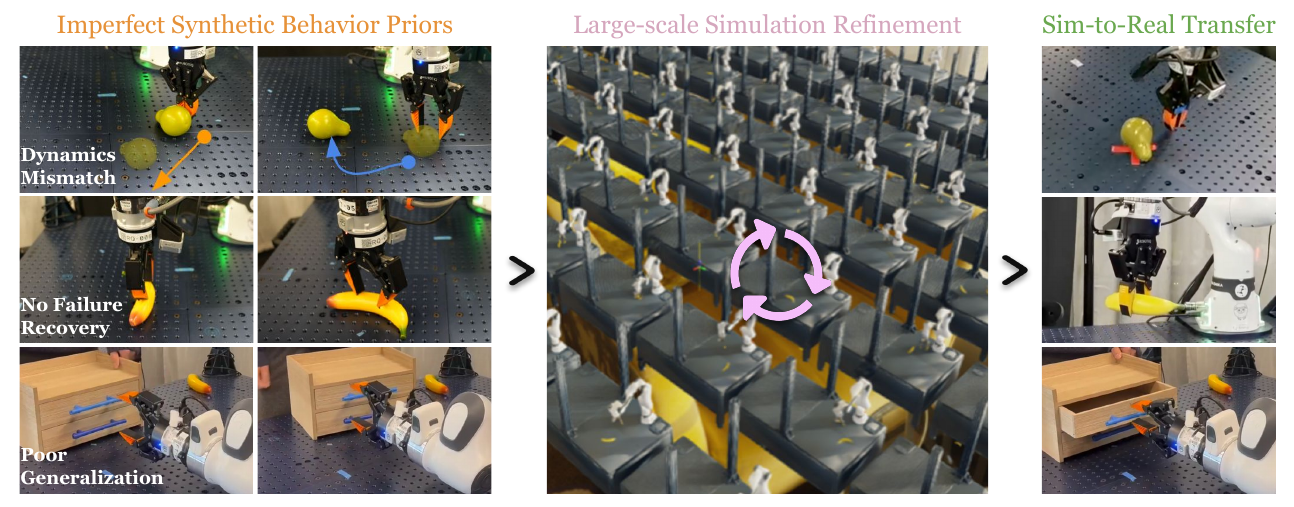}}
   \captionof{figure}{\textsc{ExpertGen} pipeline: (left) generative modeling of imperfect behavior priors; (middle) steering prior model in massively parallel simulation using reinforcement learning; and (right) visual distillation in simulation with DAgger for zero-shot sim-to-real transfer.
   }
\label{fig:teaser}
\end{center}%

\begin{abstract}
Learning generalizable and robust behavior cloning policies requires large volumes of high-quality robotics data. While human demonstrations (e.g., through teleoperation) serve as the standard source for expert behaviors, acquiring such data at scale in the real world is prohibitively expensive. This paper introduces \textsc{ExpertGen}, a framework that automates expert policy learning in simulation without any human demonstrations, while preserving strong sim-to-real transfer performance. \textsc{ExpertGen} first initializes a behavior prior in using a diffusion policy trained on synthetic data (e.g., Code as Policy). Reinforcement learning is then used to steer this prior \emph{in massively parallel simulation} toward high task success by optimizing the diffusion model’s initial noise while keeping the denosing process frozen. In this way, \textsc{ExpertGen} regularizes exploration to remain within safe behavior manifolds, enabling stable learning from sparse rewards without additional reward engineering. Extensive experiments on challenging manipulation benchmarks demonstrate that \textsc{ExpertGen} reliably acquires high-quality policies with strong robustness and generalization. On long-horizon manipulation tasks, \textsc{ExpertGen} achieves success rates of up to 85\%, while on industrial assembly benchmarks it achieves a 91.1\% overall success rate, outperforming all baseline methods. To validate sim-to-real transfer, the learned state-based expert policies are further distilled into visuomotor policies via DAgger and successfully deployed on real hardware.
\end{abstract}



\section{Introduction}
Recent progress in synthetic robotics data generation has enabled scalable creation of manipulation demonstrations in simulation. Existing approaches span several paradigms. Demonstration augmentation methods, such as MimicGen~\cite{mandlekar2023mimicgen}, expand a small number of human demonstrations through demonstration perturbation, geometric retargeting, and skill composition~\cite{mandlekar2023mimicgen,garrett2024skillmimicgen,jiang2025dexmimicgen}. More recent approaches further reduce human effort by generating demonstrations from task-and-motion planning (TAMP) or Code as Policies (CaP) built on reusable motion primitives~\cite{gong2025anytask,liang2022code}. Together, these approaches significantly reduce the cost of robotics data generation.

However, simply scaling demonstration quantity does not guarantee robust learned behaviors. Existing synthetic data generation pipelines primarily generate trajectories around nominal successful executions. MimicGen-style approaches rely on heuristically designed perturbation strategies around demonstrations, while TAMP and CaP are constrained by predefined skills and planning heuristics. Consequently, these methods provide limited coverage of failure states and recovery behaviors.
In principle, reinforcement learning (RL) can address these limitations by directly optimizing closed-loop task success and discovering recovery behaviors through trial-and-errors. However, large-scale RL-based robotics data generation typically requires carefully engineered dense rewards, which substantially limits scalability across tasks and domains~\cite{huagensim2,taomaniskill3}. To address these challenges, this paper proposes to \emph{treat synthetic demonstrations as a scalable behavior prior and subsequently refines this prior through RL using only sparse task-level rewards.}

To this end, this paper introduces \textsc{ExpertGen}, a scalable framework for sim-to-real expert policy acquisition from \emph{synthetic behavior priors}. \textsc{ExpertGen} first trains a diffusion policy on demonstrations synthesized from scripted policies. These synthetic demonstrations typically encode safe and feasible behaviors while achieving non-trivial task success rates; however, they may still exhibit limited state coverage, incomplete recovery behaviors, or dynamics mismatch. \textsc{ExpertGen} then adopts diffusion steering reinforcement learning (DSRL)~\cite{wagenmaker2025steering}, which optimizes only the diffusion model’s initial noise while keeping the pretrained denoising process frozen. Building on this formulation, \textsc{ExpertGen} extends diffusion steering to massively parallel robotics simulation using an off-policy RL framework designed for high-throughput large-batch training. 
One of our key insights is that the behavior prior need only be {\em roughly} correct; as long as expert behaviors lie somewhere on the prior manifold, large-scale RL can still efficiently find such expert policies under sparse rewards.

Empirical evaluations on the \textsc{AnyTask}~\cite{gong2025anytask} and AutoMate~\cite{tang2024automate} benchmarks demonstrate that \textsc{ExpertGen} consistently improves task success rates across long-horizon manipulation and dexterous assembly tasks, outperforming alternative policy fine-tuning paradigms such as residual RL while requiring only sparse rewards. For sim-to-real transfer, the learned expert policies are distilled into visuomotor policies through large-scale DAgger~\cite{ross2011reduction} training with point-cloud or RGB observations under domain randomization, enabling robust zero-shot deployment on real robotic hardware.

The major contributions of the paper are
\begin{itemize}[itemsep=1pt, topsep=2pt, parsep=0pt, partopsep=0pt]
    \item We introduce \textsc{ExpertGen}, a scalable framework that transforms imperfect synthetic behavior priors into sim-to-real-ready expert policies.

    \item We showcase that DSRL can be effectively extended to massively parallel simulation, enabling high success, robustness, and generalization, while preserving behavior manifolds.

    \item We demonstrate robust zero-shot sim-to-real transfer of visuomotor policies distilled from expert teacher policies, including point-cloud- and RGB-based policies.
\end{itemize}
\vspace{-10pt}
\section{Related Work}

\paragraph{Offline-to-Online Reinforcement Learning.}
\vspace{-5pt}
A growing body of work studies how to refine policies learned from offline demonstrations using online interaction. Residual RL~\cite{silver2018residual} constrains exploration through residual action spaces and has shown promising results in dexterous manipulation and loco-manipulation~\cite{ankile2025imitation,zhao2025resmimic,wang2026omnixtreme}. However, these approaches are often sensitive to residual action magnitudes and exploration schedules~\cite{yuan2024policy,dodeja2025accelerating}, making them difficult to scale across diverse tasks. More recently, diffusion policies~\cite{chi2025diffusion} have motivated specialized offline-to-online refinement methods~\cite{ren2024diffusion,wagenmaker2025steering,yi2026flow,mcallister2025flow,lyu2026flow,celik2025dime}. In particular, diffusion steering reinforcement learning (DSRL)~\cite{wagenmaker2025steering} refines behavior by steering the initial diffusion noise while keeping the pretrained denoising process frozen, preserving the learned action manifold during online optimization. \textsc{ExpertGen} builds on this formulation and extends it to massively parallel simulation for scalable expert policy learning.

\paragraph{Synthetic Robotics Data Generation.}
Another line of work addresses the scalability bottleneck of robotics data collection through synthetic demonstration generation and data augmentation. Prior approaches expand small human datasets through trajectory perturbation, geometric retargeting, skill composition, or reusable motion primitives~\cite{mandlekar2023mimicgen,garrett2024skillmimicgen,jiang2025dexmimicgen,xue2025demogen,zhou2025reinforcegen}, while more recent methods synthesize demonstrations directly from scripted policies, planning systems, or large language model–generated robot programs~\cite{liang2022code,chen2025robotwin,tian2025interndata,gong2025anytask,huagensim2}. These methods substantially reduce the cost of collecting robotics demonstrations and enable scalable generation of long-horizon and dexterous manipulation data. However, because generated trajectories are typically centered around nominal successful executions, they often provide limited coverage of recovery behaviors, failure states, and out-of-distribution configurations. \textsc{ExpertGen} treats synthetic demonstrations as scalable behavior priors and subsequently refines these priors with RL using only sparse task-level rewards.

\section{Preliminaries}
\vspace{-5pt}
This section formalizes the learning problem and introduces the notation used throughout the paper.

\textbf{Markov Decision Process} \quad The manipulation tasks are formulated as a Markov decision process (MDP), defined by the tuple$\mathcal{M}=(\mathcal{S}, \mathcal{A}, \mathcal{T}, r, \gamma)$, where $\mathcal{S}$ is the state space, $\mathcal{A}$ is the action space,
$\mathcal{T}(s_{t+1}\mid s_t,a_t)$ is the transition function,
$r:\mathcal{S}\times\mathcal{A}\rightarrow\mathbb{R}$ is the reward function,
and $\gamma\in(0,1]$ is the discount factor.
The objective is to learn an expert policy $\pi_E$ that maximizes the expected discounted return:
$
\max_{\pi}
\quad
\mathbb{E}_{\tau \sim \pi}
\left[
\sum_{t=0}^{T}
\gamma^t r(s_t,a_t)
\right]
$, 
where
$\tau=(s_0,a_0,s_1,a_1,\ldots,s_T,a_T)$
denotes a trajectory induced by the policy interacting with the environment.

In this work, the reward function is defined as a sparse binary task success signal. Let $\mathbb{I}_{\mathrm{succ}}(s_t)$ denote an indicator function that evaluates to $1$ if the task is successfully completed at state $s_t$ and $0$ otherwise. The reward is given by $r(s_t,a_t)=\mathbb{I}_{\mathrm{succ}}(s_t)$, 
which provides sparse supervision by assigning nonzero reward only upon successful task completion.

\textbf{Diffusion Policy} \quad This paper considers diffusion policy for all imitation learning training. Specifically, diffusion policy considers demonstrations as samples from a diffusion denoising process. Given a state $s_t$ at time step $t$, the policy represents the conditional distribution of an action chunk~\cite{zhao2023learning} $\mathbf{a}_t = a_{t:t+H}$ of length $H$ via a diffusion model~\cite{chi2025diffusion}. During training, Gaussian noise is progressively added to expert trajectories according to a forward process
$
q(\mathbf{a}_{t}^k \mid \mathbf{a}_t^0) = \mathcal{N}\!\left(\mathbf{a}_{t}^k; \sqrt{\bar{\alpha}_k}\, \mathbf{a}_t^0,\; (1-\bar{\alpha}_k) I\right)
$, 
where $\bar{\alpha}_k = \prod_{i=1}^k \alpha_i$ defines the noise schedule. The diffusion policy is trained to predict the injected noise using a denoising network $\epsilon_\theta$, by minimizing the standard diffusion objective:
\begin{equation}
\mathcal{L}_{\text{diff}} =
\mathbb{E}_{\mathbf{a}_t^0, s_t,\, k,\, \epsilon \sim \mathcal{N}(0,I)}
\left[
\left\| \epsilon - \epsilon_\theta(\mathbf{a}_{t}^k, s_t, k) \right\|^2
\right].
\end{equation}

At inference time, action trajectories are generated by iteratively applying the reverse diffusion process starting from Gaussian noise $\mathbf{a}_t^K \sim \mathcal{N}(0,I)$. To reduce denoising steps, we adopt Denoising Diffusion Implicit Models (DDIM)~\cite{song2020denoising} sampling, which uses deterministic denoising process.

\section{\textsc{ExpertGen}}
\begin{figure}
    \centering
    \includegraphics[width=0.80\linewidth]{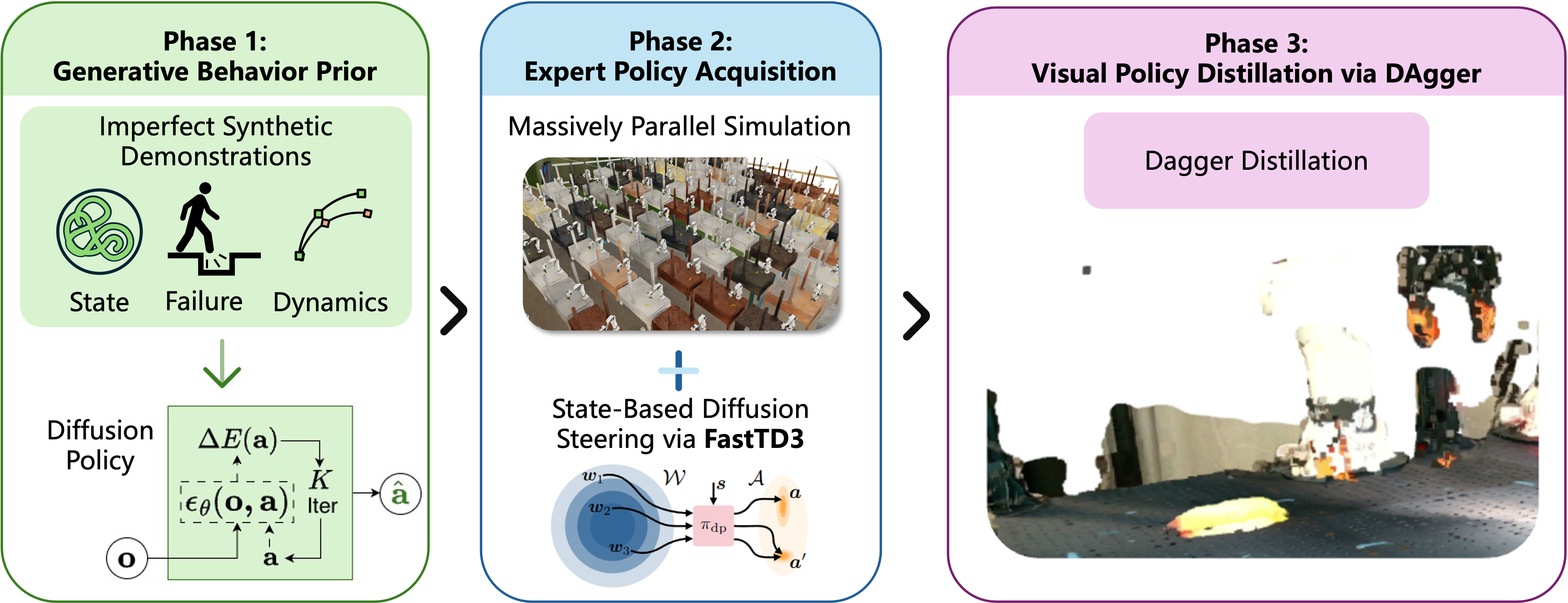}
    \caption{\textsc{ExpertGen} training pipeline: generative modeling of imperfect behavior priors using a state-based diffusion policy (Phase 1); steering diffusion policy in massively parallel simulation using FastTD3 (phase 2); visual policy distillation using DAgger from expert teachers (phase 3).}
    \vspace{-10pt}
    \label{fig:sys_diagram}
\end{figure}

This section details \textsc{ExpertGen}, a three-stage framework (Fig.~\ref{fig:sys_diagram}) for acquiring sim-to-real-ready expert policies from imperfect but scalable synthetic demonstrations. The key idea is to first learn a structured behavior prior from purely synthetic data, then improve this prior through sparse-reward reinforcement learning in simulation, and finally distill the resulting state-based expert into deployable visuomotor policies.

\subsection{Generative Behavior Prior Modeling}
\textsc{ExpertGen} assumes access to a dataset of synthetic demonstration trajectories
$
\mathcal{D}_{\mathrm{syn}}=\{\tau_i\}_{i=1}^{N}
$. Each trajectory $
\tau_i=\{(s_t,a_t)\}_{t=0}^{T}
$ is generated entirely in simulation rather than collected through real-world teleoperation. Specifically, \textsc{ExpertGen} considers synthetic data generated from both AnyTask~\cite{gong2025anytask} and SkillMimicGen~\cite{garrett2024skillmimicgen} for tabletop manipulation tasks, as well as handcrafted policies for contact-rich assembly tasks.

Although these synthetic demonstrations often exhibit safe and structured behaviors with nontrivial task success, they typically suffer from limited state coverage, weak recovery behaviors, and mismatch between the synthetic generation process and the target task distribution. Consequently, directly cloning such demonstrations does not produce robust expert policies. To model this imperfect behavior prior, \textsc{ExpertGen} trains a state-conditioned diffusion policy $\pi_{\mathrm{prior}}(\mathbf{a}_t \mid s_t)$ on $\mathcal{D}_{\mathrm{syn}}$. The diffusion policy serves as a structured generative prior that captures the motion manifold of the synthetic demonstrations while remaining amenable to RL-based steering.

\subsection{Expert Policy Acquisition}
\label{sec:design_choices}
This subsection describes how \textsc{ExpertGen} acquires task-specialized expert policies by combining diffusion steering with massively parallel simulation and off-policy RL.


\textbf{Diffusion steering reinforcement learning} \quad
Let $z_t$ denote the initial noise input to the diffusion model. A steering policy $q_{\phi}(z_t \mid s_t)$ selects this noise conditioned on the current state. The diffusion prior $\pi_{\mathrm{prior}}(\mathbf{a}_t \mid s_t)$ then maps $(s_t,z_t)$ to an action chunk. RL optimizes $q_{\phi}$ using sparse task-level rewards, while the frozen diffusion decoder constrains the resulting actions to remain close to the behavior distribution learned from synthetic data.

\textbf{Massively parallel off-policy RL} \quad
While the original DSRL formulation focuses on Soft Actor--Critic (SAC)~\cite{haarnoja2018soft}, which prioritizes sample efficiency, \textsc{ExpertGen} targets a fundamentally different regime: \emph{massively parallel, large-scale simulation training}. To support this setting, \textsc{ExpertGen} adopts an off-policy RL training inspired by FastTD3~\cite{seo2025fasttd3}, which integrates several design choices for high-throughput robotic learning in massively parallel simulation.

Specifically, \textsc{ExpertGen} incorporates several design choices. First, policy optimization is performed using large-batch updates collected from thousands of parallel environments, substantially improving training throughput and effective learning under sparse rewards. Second, a distributional critic~\cite{bellemare2017distributional} is employed to improve value estimation in long-horizon sparse-reward settings. Third, mixed Gaussian noise~\cite{li2023parallel} is introduced in the steering latent space to maintain exploration diversity across parallel environments while regularizing the steering policy to be close to the normal distribution used to train the base diffusion policy. Empirically, this mixed-noise strategy is critical for preserving the prior behavior manifold after RL steering.


\subsection{Visuomotor Policy Distillation through DAgger}
The expert policies learned in simulation operate on privileged states that are unavailable at deployment time. To obtain deployable visuomotor policies, the expert is distilled into an observation-based policy $\pi(a \mid o)$ using Dataset Aggregation (DAgger)~\cite{ross2011reduction}, where observations $o=f(s)$ consist of RGB images or point clouds. To improve sim-to-real transfer, extensive visual domain randomization is applied during distillation, including randomized textures, lighting, object appearances, and camera poses. More details are presented in Sec.~\ref{sec:dagger}.


\section{Experiment Setup}
\begin{figure}[H]
    \centering
    \vspace{-10pt}
    \includegraphics[width=0.8\textwidth]{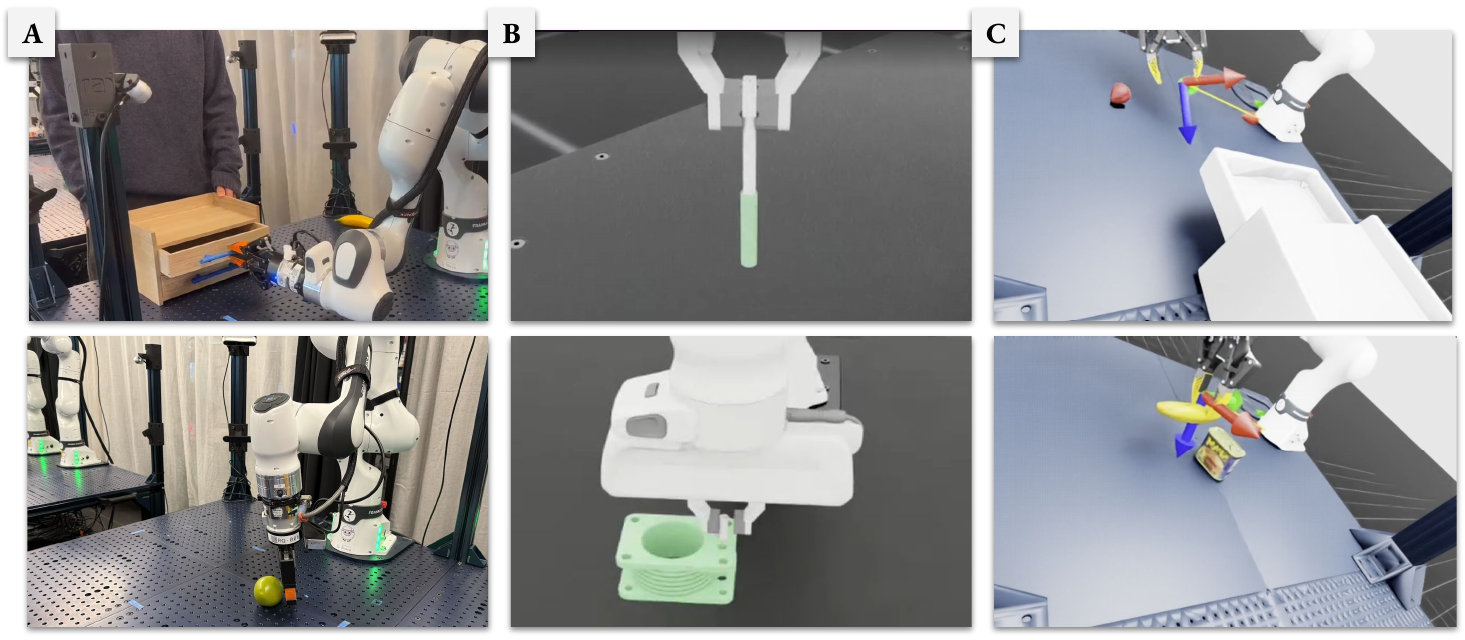}
    \caption{Illustrations of the tasks in our experiments. (A) Real-world manipulation tasks. (B) Industrial assembly tasks from AutoMate. (C) Long-horizon tasks from \textsc{AnyTask}. }
    \vspace{-10pt}
    \label{fig:benchmarks}
\end{figure}
\subsection{Benchmarks - \textsc{AnyTask}}
\textbf{Tasks} \quad \textsc{AnyTask}~\cite{gong2025anytask} is an automated framework for task design and large-scale synthetic data generation in robotic manipulation. For benchmarking, we select eight tabletop manipulation tasks spanning a broad range of behaviors, including lifting, pushing, stacking, pick-and-place, and drawer opening. These tasks cover both short-horizon and long-horizon interactions, as well as contact-rich manipulation scenarios. \autoref{fig:benchmarks} (C) visualizes two tasks from \textsc{AnyTask}.

\textbf{Synthetic data} \quad The synthetic data is synthesized by AnyTask~\cite{gong2025anytask}, which uses a LLM to program scripted policies that invokes an existing skill library. We collect 1000 demonstrations for each task from the scripted policies. The scripted polices are listed in Appendix~\ref{appendix:anytask}.

\vspace{-5pt}
\paragraph{Baselines} We compare \textsc{ExpertGen} against several representative baselines. \textbf{Scripted Policy} directly executes the LLM-synthesized scripted policies provided by \textsc{AnyTask}. \textbf{Diffusion Policy} evaluates policies trained purely via behavior cloning on synthetic demonstrations. Offline-to-Online Refinement includes Residual RL~\cite{yuan2024policy} and SMP~\cite{mu2025smp}, which both initialize from pretrained diffusion priors and improve performance through online RL. We additionally compare against the original \textbf{DSRL} formulation~\cite{wagenmaker2025steering}, which performs diffusion steering using SAC. \textbf{ExpertGen-PPO} further replaces off-policy RL with PPO to study the effect of off-policy RL in massively parallel simulation steering. Finally, \textbf{FastTD3} trains directly from sparse rewards without demonstrations or pretrained priors, measuring the difficulty of solving these tasks from scratch. More implementation details are provided in the Appendix.

\subsection{Benchmarks - AutoMate}
\textbf{Tasks}\quad AutoMate~\cite{tang2024automate} is a simulation benchmark for robotic dexterous manipulation in industrial assembly, with a particular emphasis on high-precision peg-insertion tasks. The benchmark consists of a collection of insertion scenarios, as shown in \autoref{fig:benchmarks} (B), with tight geometric tolerances, demanding accurate pose alignment and dexterous control.

\textbf{Synthetic data}\quad In the AutoMate benchmark, synthetic demonstrations are generated using a scripted disassembly policy that starts from already-assembled configurations and disassemble the parts from the sockets. Then, the disassembly trajectories are reversed and used for assembly policy learning. Because assembly and disassembly dynamics are inherently asymmetric, reversing the trajectories introduces dynamics mismatch and imperfect contact interactions during insertion. We collect 500 demonstrations for each task. Additionally, we consider two variants of the demonstrations depending on whether small noise is injected in the $x$--$y$ plane during the lifting phase of the disassembly trajectories. This perturbation increases trajectory diversity near insertion regions, producing datasets with and without $x$--$y$ noise augmentation.

\textbf{Baselines}\quad We compare against two baselines on AutoMate. \textbf{Diffusion Policy} trains diffusion policies purely via imitation learning on the reversed demonstrations, without RL refinement. We evaluate variants trained with and without the additional $x$--$y$ noise augmentation. \textbf{AutoMate} reproduces the single-task expert policies proposed in AutoMate using the original benchmark training setup.

\subsection{Evaluation Metrics}

\textbf{Success Rate} \quad
For the \textsc{AnyTask} and AutoMate benchmark, we report the success rate evaluated over 1024 and 256 rollout trajectories respectively from each baseline approach, with error bars denoting 95\% Wilson binomial confidence intervals~\cite{enwiki:1333177192}.

\textbf{Wasserstein-2 Distance} \quad
To quantify the distributional shift from the base diffusion policy, we measure the Wasserstein-2 distance between the action distributions induced by each evaluated approach and the pretrained diffusion model. For each task from \textsc{AnyTask}, we randomly sample 5 states from each of 10 demonstration trajectories, resulting in a total of 100 evaluation states. At every sampled state, we generate 512 action samples from the pretrained diffusion policy and 512 action samples from the evaluated method. The empirical Wasserstein-2 distance is then computed between the two action distributions and averaged across all sampled states. 

\section{Experimental Results}
\vspace{-10pt}
\begin{table*}[htb!]
\centering
\resizebox{\linewidth}{!}{%
\begin{tabular}{l|P{1.6cm}|P{1.6cm}|P{1.6cm}|P{1.6cm}|P{1.6cm}|P{1.6cm}|P{2.cm}|P{2.cm}}
\toprule
\multirow{3}{*}{\diagbox{Methods}{Tasks}} & \vspace{0.0pt}\centering\textit{Lift\qquad Banana} & \vspace{0.0pt}\centering\textit{Lift\qquad Brick} & \vspace{0.0pt}\centering\textit{Lift\qquad Peach} & \vspace{0.0pt}\centering\textit{Open\qquad  Drawer} & \vspace{0.0pt}\centering\textit{Push Pear \qquad to Center} & \vspace{0.0pt}\centering\textit{Stack Banana on Can} & \vspace{0.0pt}\centering\textit{Put Object In Closed Drawer} & \textit{Place Strawberry\quad In Bowl} \\
\midrule
 Scripted Policy               & 56.3 $\pm$ 3.1 & 66.4 $\pm$ 3.0 & 60.9 $\pm$ 3.1 & 28.1 $\pm$ 2.8 & 18.0 $\pm$ 2.4  & 26.6 $\pm$ 2.7  & ~~9.4 $\pm$ 1.8  & 43.0 $\pm$ 2.9   \\
 Diffusion Policy               & 69.8 $\pm$ 2.8 & 72.6 $\pm$ 2.7 & 28.5 $\pm$ 2.8  & 77.6 $\pm$ 2.6 & 50.5 $\pm$ 3.1  & 16.3 $\pm$ 2.3  & ~~0.5 $\pm$ 0.5 & ~~6.6 $\pm$ 1.5   \\
 \midrule
 Residual RL               & 84.3 $\pm$ 2.0 & 98.4 $\pm$ 0.8 & 75.9 $\pm$ 2.6 & \textbf{99.5} $\pm$ \textbf{0.5} & 56.5 $\pm$ 3.0  & ~~0.0 $\pm$ 0.2  & 13.6 $\pm$ 2.1  & ~~1.3 $\pm$ 0.7   \\
 SMP               & ~~0.0 $\pm$ 0.2 & ~~0.0 $\pm$ 0.2 & ~~0.0 $\pm$ 0.2 & \textbf{99.9 $\pm$ 0.3} & ~~2.1 $\pm$ 0.2  & ~~0.0 $\pm$ 0.2  & ~~0.0 $\pm$ 0.2  & ~~1.2 $\pm$ 0.7   \\
 DPPO & 96.0 $\pm$ 1.0 & 96.7 $\pm$ 1.1 & 95.2 $\pm$ 1.3 & 98.1 $\pm$ 0.9 & 26.7 $\pm$ 3.2 & 7.8 $\pm$ 0.8 & 1.0 $\pm$ 0.7 & 4.0 $\pm$ 1.4 \\
 \midrule
 DSRL & 72.4 $\pm$ 2.6  & 77.8 $\pm$ 2.5 & ~~1.2 $\pm$ 0.7 & 94.9 $\pm$ 1.4 & 44.8 $\pm$ 3.0 & ~~6.5 $\pm$ 1.5 & ~~1.2 $\pm$ 0.7 & ~~3.0 $\pm$ 1.2 \\
 FastTD3               & ~~0.0 $\pm$ 0.2 & ~~0.0 $\pm$ 0.2 & ~~0.0 $\pm$ 0.2 & ~~0.0 $\pm$ 0.2 & ~~0.0 $\pm$ 0.2  & ~~0.4 $\pm$ 0.4  & ~~0.0 $\pm$ 0.2  & ~~1.5 $\pm$ 0.8   \\
  ExpertGen-PPO               & 89.6 $\pm$ 1.9 & 90.0 $\pm$ 1.8 & 81.9 $\pm$ 2.4 & 96.7 $\pm$ 1.1 & 69.8 $\pm$ 2.8  & 57.0 $\pm$ 3.0  & 54.9 $\pm$ 3.0  & 44.3 $\pm$ 3.0   \\
\midrule
 ExpertGen (ours)              & \textbf{99.8 $\pm$ 0.3} & \textbf{99.7 $\pm$ 0.4} & \textbf{99.3 $\pm$ 0.0} & \textbf{~100 $\pm$ 0.1} & \textbf{83.3 $\pm$ 2.0}  & \textbf{67.2 $\pm$ 2.9}  & \textbf{80.7 $\pm$ 2.4}  & \textbf{52.1 $\pm$ 3.1}   \\

\bottomrule
\end{tabular}}
\caption{The success rates (\%) of the evaluated state-based policies on \textsc{AnyTask} benchmark. Bold number indicates the best number across all the approaches.}
\label{tab:results}
\end{table*}

\begin{figure*}
    \centering
    \includegraphics[width=1.0\linewidth]{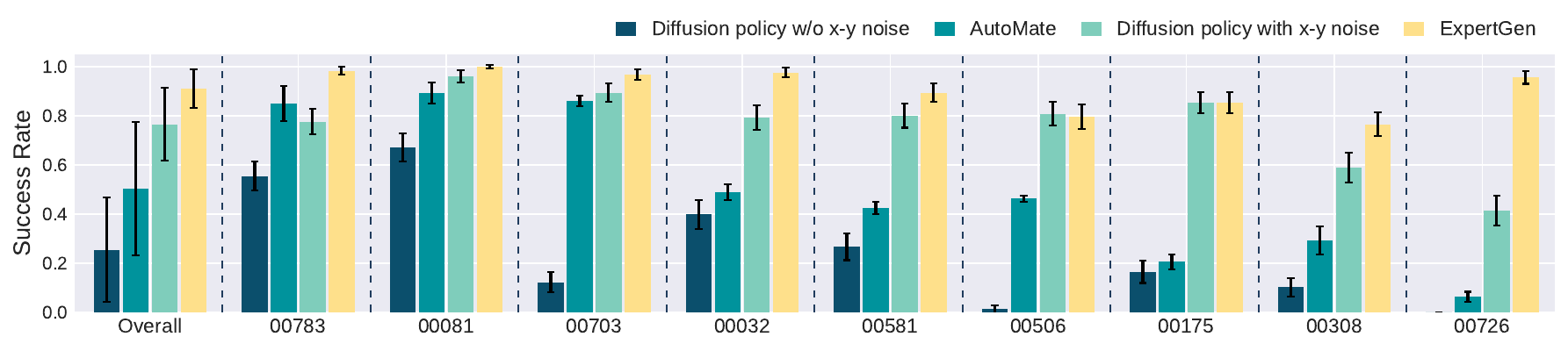}
    \caption{The success rates (\%) of the evaluated approaches on assets from the AutoMate benchmark. \textsc{ExpertGen} outperforms all other baselines with an overall success of 91.1\%. By introducing x-y noise, the diffusion policy demonstrates higher success rates compared to the no x-y noise.}
    \label{fig:automate_result}
    \vspace{-10pt}
\end{figure*}

In this section, we discuss the major results including the expert performance, recovery capacity, trajectory smoothness, and visuomotor policy performance after transferring to the real world.
\subsection{Expert Policy Acquisition}
Starting from imperfect synthetic behavior priors, \textsc{ExpertGen} consistently learns near-perfect expert policies on both benchmarks. On the \textsc{AnyTask} benchmark (\autoref{tab:results}), \textsc{ExpertGen} achieves the highest success rates across all eight evaluation tasks. For short-horizon tasks, success rates approach 100\%. Performance on long-horizon tasks is comparatively lower, which is primarily attributed to inherently unsolvable configurations in the benchmark. Failure cases include objects being placed too close to the drawer frame, causing blockage during opening. Despite these challenges, \textsc{ExpertGen} consistently outperforms all baseline methods. The strong improvement over DSRL reveals the importance of the design choices in Sec.~\ref{sec:design_choices} that support massively-parallel simulation. Beyond tabletop manipulation, as shown in \autoref{fig:automate_result}, \textsc{ExpertGen} attains an average success rate of 90.5\% across nine high-precision industrial assembly tasks in AutoMate, demonstrating its ability to solve dexterous manipulation tasks from behavior priors with inaccurate contact dynamics.

\subsection{Failure Recovery Capacity}
\begin{table*}[thb!]
    \centering
    \resizebox{0.8\linewidth}{!}{%
    \begin{tabular}{c|cc|cc|cc}
        \toprule
         \multirow{2}{*}{Methods} &  \multicolumn{2}{c|}{\emph{Lift Banana}}&  \multicolumn{2}{c|}{\emph{Open Drawer}}&  \multicolumn{2}{c}{\emph{Push Pear}}\\
        \cmidrule(lr){2-3} \cmidrule(lr){4-5} \cmidrule(lr){6-7}
         &  Open&  Force&  Open&  Force&  Open& Force\\
         \midrule
         Diffusion Policy& 72.5 (7.6$\downarrow$) & ~1.9 (79.9$\downarrow$) & 78.2 (8.7$\downarrow$) & ~58.0 (28.9$\downarrow$) &  47.1 (1.9$\downarrow$) & 15.8 (33.1$\downarrow$) \\
         Residual RL & 83.8 (1.0$\downarrow$) & 46.7 (37.5$\downarrow$) & 96.9 (2.6$\downarrow$) & \textbf{99.7 (0.2$\uparrow$)}~ & 57.1( 0.7$\uparrow$) & 38.4 (18.1$\downarrow$) \\
         \textsc{ExpertGen}& \textbf{99.4 (0.3$\downarrow$)} & \textbf{56.6 (43.1$\downarrow$)} & \textbf{99.9 (0.0$\downarrow$)} & \textbf{99.7 (0.2$\downarrow$)}~ & \textbf{85.4 (1.2$\uparrow$)} & \textbf{38.7 (45.3$\downarrow$)} \\
         \bottomrule
    \end{tabular}}
    \caption{Success rates (\%) with two perturbations: random gripper opening and random external force applied to the end-effector. Number in parentheses indicate the performance drop ($\downarrow$) / increase ($\uparrow$) compared to no perturbation. Bold number indicates the best success rates.}
    \label{tab:robustness}
    \vspace{-5pt}
\end{table*}
To evaluate failure recovery capacity, we introduce targeted perturbations to tasks from \textsc{AnyTask}, including random gripper opening events and random external forces applied to the end-effector in the $x$–$y$ plane. Under these perturbations, \textsc{ExpertGen} exhibits only minor performance degradation (\autoref{tab:robustness}), with average success-rate drops of 0.5\% and 28.6\% across three tasks for gripper-opening and external-force perturbations, respectively. In contrast, the base diffusion policy is highly sensitive to such disturbances, with success rates dropping to nearly zero when random external forces are applied. Residual RL suffers from similar amount of performance drop compared to \textsc{ExpertGen}, because they are both trained with RL under perturbation. However, \textsc{ExpertGen} consistently showcases higher success rate than Residual RL even with the perturbation. These results demonstrate that \textsc{ExpertGen} policies possess strong failure recovery capabilities and can provide high-quality supervision for DAgger distillation.

\subsection{Preserving the Action Manifold}
\vspace{-5pt}
\begin{table*}[htb!]
\centering
\resizebox{\linewidth}{!}{%

\begin{tabular}
{l|C{1.8cm}|C{1.8cm}|C{1.8cm}|C{1.8cm}|C{1.8cm}|C{1.8cm}|C{2.cm}|C{2.cm}}
\toprule
\multirow{3}{*}{\diagbox{Methods}{Tasks}} & \vspace{0.0pt}\centering\textit{Lift \\ Banana} & \vspace{0.0pt}\centering\textit{Lift \\ Brick} & \vspace{0.0pt}\vspace{0.0pt}\centering\textit{Lift \\ Peach} & \vspace{0.0pt}\centering\textit{Open \\  Drawer} & \vspace{0.0pt}\centering\textit{Push Pear \\ to Center} & \centering\textit{Stack Banana on Can} & \centering\textit{Put Object In Closed Drawer} & \textit{Place Strawberry\quad In Bowl} \\
\midrule
Residual RL & 0.23 $\pm$ 0.04 & 0.22 $\pm$ 0.02 & 0.20 $\pm$ 0.02 & 0.22 $\pm$ 0.02 & 0.23 $\pm$ 0.01 & \textbf{0.23 $\pm$ 0.02} & \textbf{0.24 $\pm$ 0.01} & \textbf{0.20 $\pm$ 0.02} \\
 ExpertGen (ours) & \textbf{0.08 $\pm$ 0.03} & \textbf{0.15 $\pm$ 0.04} & \textbf{0.04 $\pm$ 0.04} & \textbf{0.09 $\pm$ 0.11} & \textbf{0.14 $\pm$ 0.05} & \textbf{0.38 $\pm$ 0.15} & 0.39 $\pm$ 0.12 & \textbf{0.18 $\pm$ 0.08}  \\
 \bottomrule
\end{tabular}}
\caption{The Wasserstein-2 distance between the evaluated policies and base diffusion policies.
}
\vspace{-5pt}
\label{tab:naturalness}
\end{table*}
Before sim-to-real transfer, \autoref{tab:naturalness} evaluates how well different fine-tuning approaches preserve the pretrained diffusion-policy action manifold. Specifically, we measure the Wasserstein-2 distance between the action distributions generated by the learned policy and the base diffusion policy.

We train Residual RL with a relatively small residual action magnitude (0.02), as larger residual actions substantially reduce the initial success rate and lead to unstable learning. Despite remaining closer to the behavior prior, Residual RL still exhibits higher Wasserstein-2 distances on short-horizon tasks and fails to achieve meaningful performance gains on long-horizon tasks, where success rates remain near zero. In contrast, \textsc{ExpertGen} consistently achieves among the lowest Wasserstein-2 distances across tasks while attaining substantially higher task success rates, demonstrating that diffusion steering effectively preserves the pretrained action manifold while enabling effective policy improvement through RL.

\subsection{Sim-to-Real Transfer of the Visuomotor Policies}
\label{sec:dagger}
\begin{table*}[thb!]
\centering
\small
\vspace{-5pt}
\resizebox{0.9\linewidth}{!}{%
\begin{tabular}{llccccc}
\toprule
\multirow{1}{*}{\textbf{Observation}}
& \multirow{1}{*}{\textbf{Method}}
& \makecell{\textbf{Number of} \textbf{Demos.}}
& \makecell{\textit{Lift} \textit{Banana}}
& \makecell{\textit{Lift} \textit{Brick}}
& \makecell{\textit{Push Pear} \textit{to Center}}
& \makecell{\textit{Open} \textit{Drawer}} \\
\midrule

\multirow{2}{*}{Point-cloud}
& \textsc{ExpertGen} & DAgger & 75.0\% & -- & 65.0\% & 85.0\% \\
& \textsc{AnyTask}~\cite{gong2025anytask} & 20k & 73.3\% & -- & 16.7\% & 42.5\% \\

\midrule

\multirow{2}{*}{RGB}
& \textsc{ExpertGen} & 20k & 80.0\% & 65.0\% & 45.0\% & 65.0\% \\
& \textsc{AnyTask}~\cite{gong2025anytask} & 20k & 80.0\% & 55.0\% & 35.0\% & 10.0\% \\

\bottomrule
\end{tabular}}
\caption{Real-world evaluation results of  visuomotor policies. Policies are trained from either \textsc{ExpertGen} or \textsc{AnyTask}~\cite{gong2025anytask} scripted-policy rollouts. While \textsc{AnyTask} achieve reasonable performance on simple pick tasks, they struggle in contact-rich manipulation, highlighting the importance of \textsc{ExpertGen}'s RL-refined teacher policies for robust sim-to-real transfer.}
\label{tab:real_world_eval}
\vspace{-10pt}
\end{table*}


In this section, we present real-world results for \textsc{ExpertGen} on selected tasks from \textsc{AnyTask}. Details of the real-world setup can be found in the Appendix~\ref{appendix:real}.



\textbf{Real-world transfer of visuomotor policies} \quad
\textsc{ExpertGen} teacher policies are distilled into point-cloud-based policies using DAgger and into RGB-based policies using standard behavioral cloning (BC). These policies are compared against counterparts trained from \textsc{AnyTask}~\cite{gong2025anytask} scripted-policy demonstrations. For RGB policies, training uses 20k offline demonstrations with extensive visual domain randomization~\cite{singh2025dextrahrgbvisuomotorpoliciesgrasp,yin2026emergentdexteritydiverseresets}, including randomized camera extrinsics, dome light textures, object textures, table textures, and HSV jitter. The RGB policy takes $224 \times 224$ RGB images together with end-effector poses as input. All methods use the same policy architecture and domain randomization strategy to ensure a fair comparison. As shown in \autoref{tab:real_world_eval}, while both approaches achieve comparable performance on simple pick-and-place tasks, \textsc{ExpertGen} consistently outperforms the \textsc{AnyTask} on tasks involving articulation or complex contact dynamics, such as \textit{Push Pear to Center} and \textit{Open Drawer}. These results suggest that RL-refined teacher policies are critical for providing supervision for complex interactions that are difficult to capture from scripted policies.

\textbf{DAgger distillation} \quad
To verify the necessity of DAgger training, we ablate the schedule of the teacher rollout ratio $\alpha$ on \textit{Lift Banana} task. Specifically, $\alpha$ is linearly annealed from 1.0 to a final value between 50k and 100k global training steps, and training continues until 200k steps. While simulation success rates remain similar across different schedules, real-world performance exhibits a clear sweet spot at a low, non-zero final $\alpha$. The highest success rate of 75.0\% is achieved with a final $\alpha=0.2$, compared to 65.0\% for pure student training ($\alpha=0.0$), 63.3\% for $\alpha=0.4$, and 55.0\% for maintaining full teacher influence ($\alpha=1.0$). These results suggest that retaining a small amount of teacher correction improves robustness to real-world noise, while both fully removing the teacher and excessively relying on teacher rollouts degrade performance.

\section{Conclusion and Limitations}
This paper introduced \textsc{ExpertGen}, a scalable framework for sim-to-real expert policy acquisition from imperfect synthetic behavior priors. \textsc{ExpertGen} first learns a structured diffusion-policy prior from synthetic demonstrations and subsequently refines this prior through diffusion steering reinforcement learning in massively parallel simulation using only sparse task-level rewards. Across long-horizon manipulation and dexterous assembly benchmarks, \textsc{ExpertGen} consistently improves robustness, recovery behaviors, and task success rates compared to the pretrained policy while preserving the action manifold. Furthermore, the resulting expert policies can be distilled into visuomotor policies through DAgger and transferred zero-shot to real robotic hardware. These results indicate that \emph{learning from imperfect synthetic priors and refining them through large-scale sim-to-real RL is a scalable strategy for robotics expert policy acquisition}. The main limitation of \textsc{ExpertGen} lies in that the performance is bounded by the quality and diversity of the synthetic behavior priors. While RL steering can substantially improve robustness and recovery behaviors, extremely poor or highly biased priors may limit the performance and exploration coverage.

\clearpage
\acknowledgments{If a paper is accepted, the final camera-ready version will (and probably should) include acknowledgments. All acknowledgments go at the end of the paper, including thanks to reviewers who gave useful comments, to colleagues who contributed to the ideas, and to funding agencies and corporate sponsors that provided financial support.}


\bibliography{references.bib}  

\clearpage

\appendix

\section{Additional Analysis}

\subsection{Imperfect Synthetic Behavior Priors}
\begin{wrapfigure}{r}{0.62\columnwidth}
    \vspace{-1.0em}
    \centering

    \includegraphics[width=0.435\linewidth]{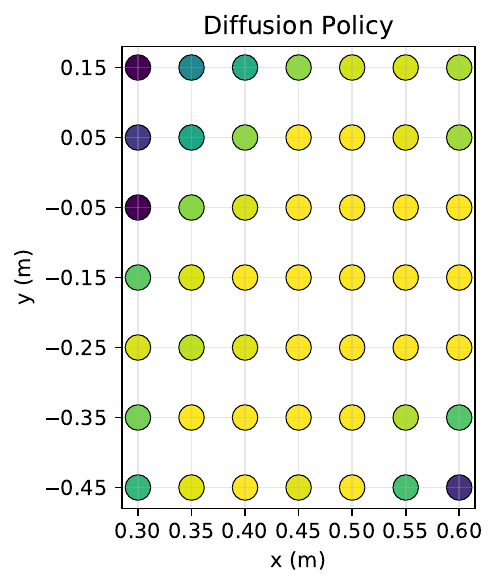}
    \hfill
    \includegraphics[width=0.555\linewidth]{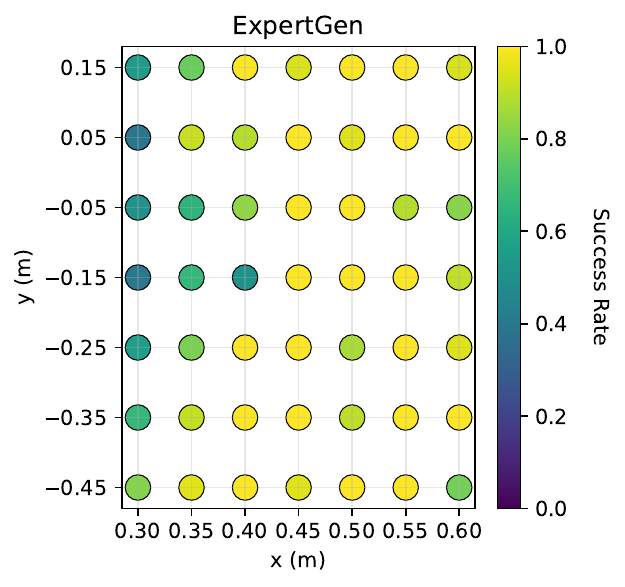}

    \caption{
    Success rate (\%) distribution of imperfect prior diffusion policy (left) and \textsc{ExpertGen} (right) over different initial object configurations for \textit{Push Pear to Center}.
    }
    \label{fig:state_coverage}
    \vspace{-1.0em}
\end{wrapfigure}

The scripted policies provide incomplete state coverage, which directly limits the quality of the learned diffusion policy. Consequently, the resulting behavior priors are imperfect and exhibit low success rates for certain initial configurations, as shown in \autoref{fig:state_coverage}. This trend is also consistent on the AutoMate benchmark, where the diffusion policy without added $x$--$y$ noise achieves the lowest success rates among all evaluated methods. Injecting $x$--$y$ noise improves spatial coverage and partially mitigates this issue; however, the resulting policy remains suboptimal. Furthermore, \autoref{tab:naturalness} shows that the diffusion policy lacks strong failure recovery capability, as evidenced by the substantial performance degradation under external perturbations.

To further isolate the impact of imperfect priors, we conduct two controlled experiments summarized in \autoref{tab:prior_deficiencies}. First, to study limited state coverage, the \textit{Lift Banana} task is trained using demonstrations in which the initial banana pose is restricted to only the left half of the workspace. Under this setting, the base diffusion policy experiences a substantial performance degradation outside the demonstrated region, whereas \textsc{ExpertGen} is able to partially extrapolate beyond the demonstrated initial-state distribution, achieving over 50\% success. Second, to evaluate robustness to dynamics mismatch, the prismatic drawer-joint friction in the \textit{Open Drawer} task is increased from 0.2 to 0.5, requiring a firmer grasp during drawer opening. While this modification significantly degrades the base diffusion policy, \textsc{ExpertGen} exhibits only negligible performance drop after steering. This robustness to dynamics mismatch is consistent with the AutoMate experiments, where priors generated by reversing disassembly trajectories contain substantial dynamics discrepancies relative to the assembly process. Overall, these results demonstrate that \textsc{ExpertGen} can effectively refine imperfect synthetic behavior priors by improving spatial generalization, robustness, and recovery behaviors through reinforcement learning.
\begin{table*}[thb!]
\centering
\begin{minipage}{0.5\linewidth}
\caption{Success rates under prior deficiencies. Arrows indicate performance drops compared to the setting without prior deficiency.}
\label{tab:prior_deficiencies}
\end{minipage}
\hfill
\begin{minipage}{0.48\linewidth}
\centering
\resizebox{\linewidth}{!}{%
\begin{tabular}{lcc}
\toprule
\textbf{Deficiencies} & \textbf{Base Policy} & \textbf{\textsc{ExpertGen}} \\
\midrule
Limited state coverage & 39.2 \; (30.6$\downarrow$) & 76.8 \; (23.2$\downarrow$) \\
Joint dynamics mismatch & 31.4 \; (41.2$\downarrow$) & 99.2 \; (0.1$\downarrow$) \\
\bottomrule
\end{tabular}
}
\end{minipage}
\end{table*}

\subsection{Number of Demonstrations}
\begin{figure}[hbt!]
    \centering
    \includegraphics[width=0.49\linewidth]{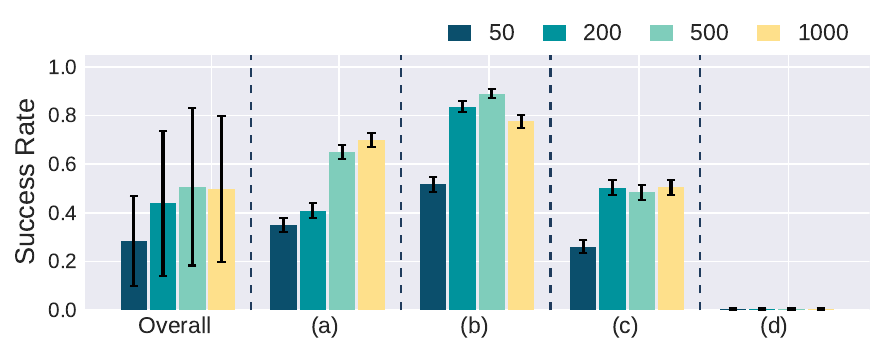}
    \includegraphics[width=0.49\linewidth]{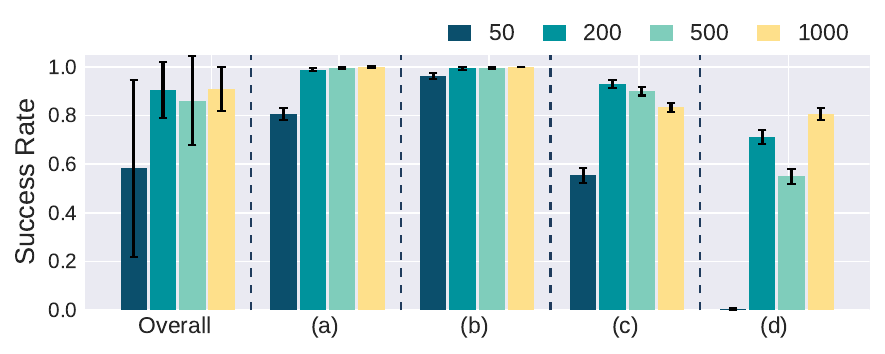}
    \caption{Success rates of base diffusion policies (left) and the \textsc{ExpertGen} policies (right) trained with \textcolor[HTML]{0B4F6C}{50}, \textcolor[HTML]{00939c}{200}, \textcolor[HTML]{7fcdbb}{500}, and \textcolor[HTML]{fee08b}{1000} imperfect demonstrations on (a) \textit{Lift Banana}, (b) \textit{Open Drawer}, (c) \textit{Push Pear to Center}, and (d) \textit{Put Object in Closed Drawer}.}
    \label{fig:num_demo}
    \vspace{-1em}
\end{figure}
We further examine how many demonstrations are required to learn a behavior prior that is sufficiently diverse to support effective refinement in simulation. Figure~\ref{fig:num_demo} reports the success rates of both the base diffusion policies and the \textsc{ExpertGen} policies trained with 50, 200, 500, and 1000 imperfect demonstrations on four \textsc{AnyTask} tasks. Performance remains largely unchanged once the dataset contains more than 200 demonstrations. A noticeable performance drop occurs only when the dataset is reduced to 50 demonstrations, where the average success rate decreases to 58.2\%. These results suggest that \textsc{ExpertGen} does not require large-scale demonstration datasets; instead, a relatively small number of imperfect demonstrations (approximately 200) is sufficient to initialize a behavior prior that can be effectively refined through reinforcement learning.

\subsection{Choice of RL algorithms}
We adopt FastTD3 as the underlying RL algorithm for expert policy acquisition. As shown in \autoref{tab:results}, while ExpertGen-PPO achieves reasonable success rates on several tasks, it consistently underperforms FastTD3 across all eight tasks in the \textsc{AnyTask} benchmark. FastTD3’s off-policy formulation enables more efficient reuse of expensive action-chunked rollouts and more stable optimization under sparse, task-level rewards, both of which are essential to the success of \textsc{ExpertGen}.

In addition, preliminary experiments with the original DSRL formulation based on Soft Actor--Critic (SAC) exhibit substantially lower asymptotic performance under the extensive domain randomization and perturbation settings used in \textsc{ExpertGen}.

\subsection{Human Motions as Behavior Priors}
While the main results use scripted policies for generating synthetic data, human demonstrations may provide richer motion diversity and more adaptive behaviors. To evaluate this hypothesis, we apply SkillMimicGen~\cite{garrett2024skillmimicgen} to \textit{Stack Banana on Can} task, on which \textsc{ExpertGen} trained with scripted policies achieves a relatively low success rate (67\%). Specifically, six human teleoperation trajectories are collected in IsaacLab and subsequently augmented to 1,000 demonstrations using SkillMimicGen. A single-task diffusion policy is then trained on this dataset, while all other components of the \textsc{ExpertGen} pipeline remain unchanged. This variant is referred to as \textsc{ExpertGen} w/ SkillMimicGen.

\begin{figure}[thb!]
    \centering

    \begin{minipage}{0.35\linewidth}
        \centering
        \includegraphics[width=\linewidth]
        {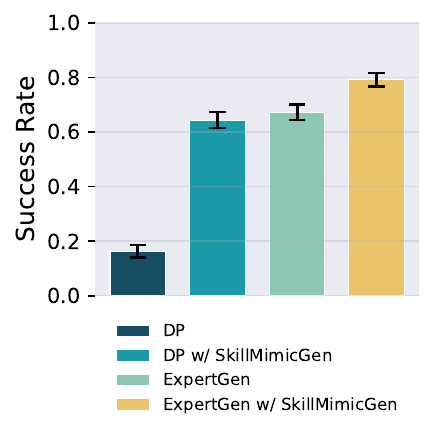}
    \end{minipage}
    \qquad
    \begin{minipage}{0.44\linewidth}
        \caption{
        Success rates (\%) of the evaluated state-based policies on the \textit{Stack Banana on Can} task. Diffusion Policy w/ SkillMimicGen trains a diffusion policy on human demonstrations augmented with SkillMimicGen~\cite{garrett2024skillmimicgen}, while \textsc{ExpertGen} w/ SkillMimicGen further refines the diffusion policy trained on SkillMimicGen data. Using human motions as behavior priors significantly improves the performance of both the diffusion policy and \textsc{ExpertGen}.
        }
        \label{fig:skillgen}
    \end{minipage}

\end{figure}

The results, summarized in \autoref{fig:skillgen}, show that both the diffusion policy and \textsc{ExpertGen} trained with SkillMimicGen data substantially outperform their counterparts trained using scripted policy demonstrations. These results suggest that human motion priors provide more diverse and adaptable behavioral patterns, which improve both imitation learning and subsequent RL refinement.

\section{Hyper-parameters}
\subsection{Diffusion Policy}
This paper considers diffusion policy for all imitation learning training. Specifically, diffusion policy considers demonstrations as samples from a diffusion denoising process. Given a state $s_t$ at time step $t$, the policy represents the conditional distribution of an action chunk~\cite{zhao2023learning} $\mathbf{a}_t = a_{t:t+H}$ of length $H$ via a diffusion model~\cite{chi2025diffusion}. During training, Gaussian noise is progressively added to expert trajectories according to a forward process
\begin{equation}
q(\mathbf{a}_{t}^k \mid \mathbf{a}_t^0) = \mathcal{N}\!\left(\mathbf{a}_{t}^k; \sqrt{\bar{\alpha}_k}\, \mathbf{a}_t^0,\; (1-\bar{\alpha}_k) I\right),
\end{equation}
where $\bar{\alpha}_k = \prod_{i=1}^k \alpha_i$ defines the noise schedule. The diffusion policy is trained to predict the injected noise using a denoising network $\epsilon_\theta$, by minimizing the standard diffusion objective:
\begin{equation}
\mathcal{L}_{\text{diff}} =
\mathbb{E}_{\mathbf{a}_t^0, s_t,\, k,\, \epsilon \sim \mathcal{N}(0,I)}
\left[
\left\| \epsilon - \epsilon_\theta(\mathbf{a}_{t}^k, s_t, k) \right\|^2
\right].
\end{equation}

At inference time, action trajectories are generated by iteratively applying the reverse diffusion process starting from Gaussian noise $\mathbf{a}_t^K \sim \mathcal{N}(0,I)$. To reduce denoising steps, we adopt Denoising Diffusion Implicit Models (DDIM)~\cite{song2020denoising} sampling, which replaces the stochastic reverse process with a deterministic update:
\begin{equation}
\mathbf{a}^{k-1}_t =
\sqrt{\bar{\alpha}_{k-1}}\, \hat{\mathbf{a}}_t^0
+ \sqrt{1-\bar{\alpha}_{k-1}}\, \epsilon_\theta(\mathbf{a}_t^k, s_{t}, k),
\end{equation}
where $\hat{\mathbf{a}}_t^0$ denotes the prediction of the clean action chunk.

\subsection{\textsc{ExpertGen}}
The set of hyper-parameters used in this paper is shown in Table \ref{tab:hyper-expertgen}
\begin{table*}[htb!]
    \centering
    \begin{tabular}{l|c c}
    \toprule
        Hyperparameters &  \textsc{AnyTask} & AutoMate\\
    \midrule
    \multicolumn{3}{c}{Generative Behavior Prior Modeling} \\
    \midrule
        Architecture & U-Net~\cite{ronneberger2015u} & U-Net \\
        Number of groups & 8 & 8 \\
        Kernel size & 5 & 5 \\
        Number of channels & [128, 256] & [128, 256] \\
        Diffusion step embedding dim. & 16 & 16 \\
        Condition type & FiLM & FiLM \\
        Action chunk length & 16 & 8 \\
        Number of epochs & 500 & 500 \\
        Batch size  & 256 & 256 \\
        Number of demonstrations & 1000 & 500 \\ 
        \midrule
        \multicolumn{3}{c}{Expert Policy Acquisition} \\
        \midrule
        Receding horizon & 8 & 8 \\
        Critic learning rate & 3e-4 & 3e-4 \\
        Actor learning rate & 3e-4 & 3e-4 \\
        Buffer size & 8196 $\times$ 1024 & 8196 $\times$ 256 \\
        Batch size & 8196 & 8196 \\
        Policy noise & 0.001 & 0.001 \\
        std\_min & 0.005 & 0.005 \\
        std\_max & 0.4 & 0.4 \\
        Number of updates & 8 & 8 \\
        Number of steps & 4 & 4 \\
        Number of atoms & 101 & 101 \\
        v\_min & 0.0 & 0.0 \\
        v\_max & 100.0 & 400.0 \\
        Discount factor & 0.99 & 0.99 \\
    \bottomrule
    \end{tabular}
    \caption{\textsc{ExpertGen} hyperparameters. The right two columns indicate the hyerparameters for \textsc{AnyTask} and AutoMate, respectively.}
    \label{tab:hyper-expertgen}
\end{table*}

\subsection{\textsc{ExpertGen}-PPO}
\textsc{ExpertGen}-PPO uses PPO for expert policy Acquisition instead of FastTD3. We list the PPO related hyper-parameters in Table \ref{tab:hyper-expertgen-ppo}. We found it important to use a large initial log-standard deviation (e.g., 0.0) to matches with the training initial noise distribution to maintain a relative high initial task success.

\begin{table*}[htb!]
    \centering
    \begin{tabular}{l|c}
    \toprule
        Hyperparameters &  \textsc{AnyTask}\\
        \midrule
        \multicolumn{2}{c}{Expert Policy Acquisition} \\
        \midrule
        Receding horizon & 8\\
        Number of steps & 64 \\
        Learning rate & 5e-4  \\
        Learning rate schedule & fixed \\
        Discount factor & 0.995\\
        e\_clip & 0.2 \\
        Entropy coefficent & -2e-4 \\
        Initial logstd & 0.0 \\
    \bottomrule
    \end{tabular}
    \caption{\textsc{ExpertGen}-PPO hyperparameters.}
    \label{tab:hyper-expertgen-ppo}
\end{table*}

\subsection{Residual RL}
Residual RL augments a fixed base prior policy $\pi_P$ with a learnable residual policy $\pi_\theta$ that predicts corrective actions. Given a state $s_t$, the base policy outputs a nominal action $a_t^0 = \pi_0(s_t)$,
while the residual policy predicts an additive correction conditioned on both the state and the base action $\delta a_t = \pi_\theta(s_t, a_t^0)$. The final action executed in the environment is given by $a_t = a_t^0 + \delta a_t$.

The critic evaluates the composed action under the environment dynamics. The state–action value function is defined as
\begin{equation}
Q_\phi(s_t, a_t^0, \delta a_t) \;\triangleq\; Q_\phi(s_t, a_t),
\end{equation}
and is trained using standard temporal-difference learning with targets computed from the combined action at the next state.

The residual actor is optimized to maximize the critic while keeping the base policy fixed:
\begin{equation}
\max_{\theta} \;
\mathbb{E}_{s_t \sim \mathcal{D}}
\left[
Q_\phi\!\left(
s_t,\,
\pi_P(s_t),\,
\pi_\theta\!\left(s_t, \pi_P(s_t)\right)
\right)
\right].
\end{equation}

We also adapt FastTD for learning the residual policy $\pi_{\theta}$. The list of hyper-parameters used for residual RL is shown in Table \ref{tab:hyper-residual_rl}.

\begin{table*}[htb!]
    \centering
    \begin{tabular}{l|c}
    \toprule
        Hyperparameters &  \textsc{AnyTask}\\
        \midrule
        \multicolumn{2}{c}{Residual RL} \\
        \midrule
        Warm-start steps & 10240 \\
        Action scale & 0.02 \\
        Receding horizon & 8 \\
        \midrule
        \multicolumn{2}{c}{FastTD3} \\
        \midrule
        Critic learning rate & 3e-4 \\
        Actor learning rate & 3e-4 \\
        Buffer size & 8196 $\times$ 1024 \\
        Batch size & 8196 \\
        Policy noise & 0.001 \\
        std\_min & 0.005 \\
        std\_max & 0.4 \\
        Number of updates & 8  \\
        Number of steps & 4  \\
        Number of atoms & 101 \\
        v\_min & 0.0 \\
        v\_max & 100.0 \\
        Discount factor & 0.995 \\
    \bottomrule
    \end{tabular}
    \caption{Residual RL hyperparameters.}
    \label{tab:hyper-residual_rl}
\end{table*}

\subsection{Score-Matching Motion Priors (SMP)~\cite{mu2025smp}}
Given a pretrained diffusion policy $\epsilon_\theta$ that models the distribution of expert action chunks, Score-Matching Motion Priors (SMP) repurpose this frozen diffusion model as a task-agnostic motion prior via score distillation sampling (SDS). During policy learning, an action chunk $\tilde{\mathbf{a}}_t^0$ generated by the current policy is evaluated under the diffusion prior by injecting Gaussian noise $\epsilon \sim \mathcal{N}(0,I)$ using the same forward diffusion process,
\begin{equation}
\tilde{\mathbf{a}}_t^k
=
\sqrt{\bar{\alpha}_k}\, \tilde{\mathbf{a}}_t^0
+
\sqrt{1-\bar{\alpha}_k}\, \epsilon .
\end{equation}
The denoising network then predicts the injected noise as
$\hat{\epsilon} = \epsilon_\theta(\tilde{\mathbf{a}}_t^k, s_t, k)$.
The squared error $\lVert \hat{\epsilon} - \epsilon \rVert_2^2$ provides a score-matching signal that measures how well the policy-generated action aligns with the expert demonstration distribution captured by the diffusion model.

To integrate this signal into reinforcement learning, SMP converts the SDS loss into a bounded prior reward,
\begin{equation}
r_{\text{smp}}
=
\exp\!\left(
- w_s \, \lVert \hat{\epsilon} - \epsilon \rVert_2^2
\right),
\end{equation}
where $w_s$ controls the strength of the motion prior. To reduce variance arising from a single diffusion timestep, the SDS loss is evaluated over a fixed set of diffusion steps and averaged before computing the reward. This prior reward is combined with a task-specific reward to train the policy using standard reinforcement learning algorithms, while keeping the diffusion model fixed throughout training. As a result, SMP provides a reusable and stationary behavior regularizer that encourages policy-generated actions to remain in-distribution with respect to expert demonstrations, without requiring adversarial training or access to the original dataset during policy optimization.

To maintain fair comparison, SMP also employs FastTD3 as the underlying RL algorithm. Table~\ref{tab:hyper-smp} lists all the hyper-parameters for our SMP implementation.

\begin{table*}[htb!]
    \centering
    \begin{tabular}{l|c}
    \toprule
        Hyperparameters &  \textsc{AnyTask}\\
        \midrule
        Warm-start steps & 10240 \\
        Action scale & 0.02 \\
        Receding horizon & 8 \\
    \bottomrule
    \end{tabular}
    \caption{SMP hyperparameters.}
    \label{tab:hyper-smp}
\end{table*}

\section{Benchmarks}

\subsection{\textsc{AnyTask}}
\label{appendix:anytask}
\begin{figure}
    \centering
    \includegraphics[width=0.4\linewidth]{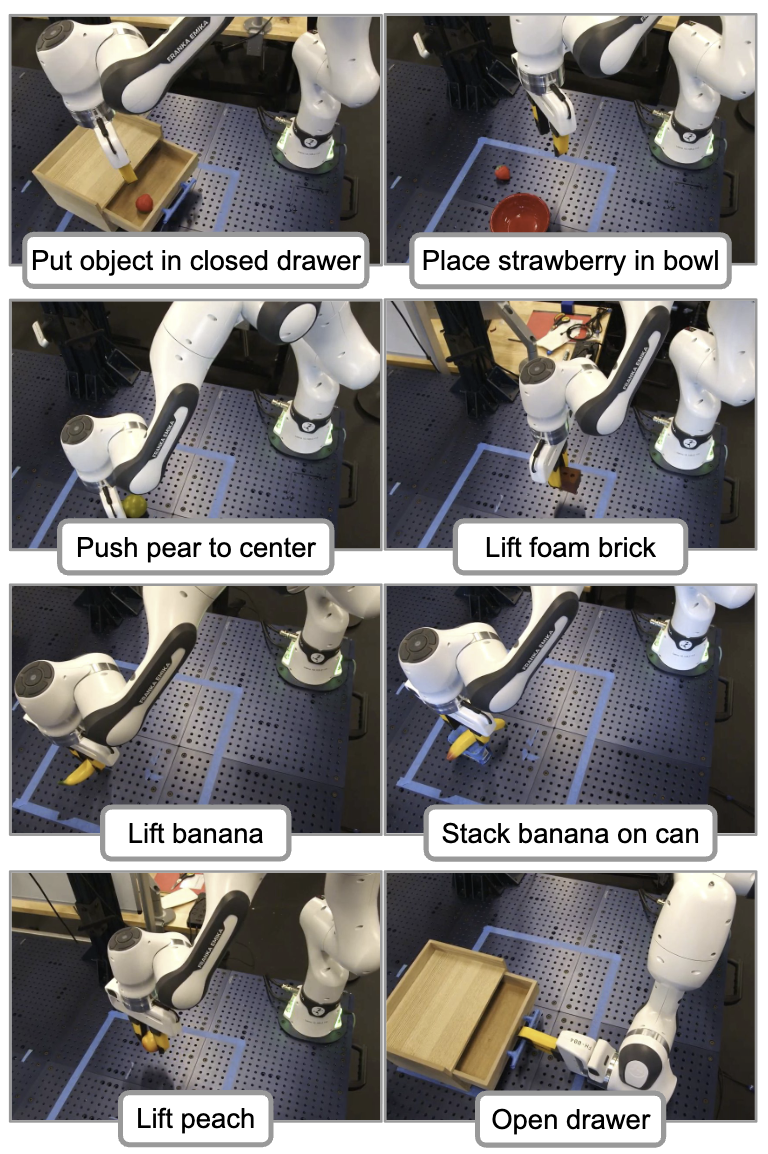}
    \caption{An overview of the selected eight tasks from \textsc{AnyTask} benchmark.}
    \label{fig:anytask_visual}
\end{figure}
\textsc{AnyTask}~\cite{gong2025anytask} is an automated framework for automatic task design and large-scale synthetic data generation in robotic manipulation. For benchmarking, eight tabletop manipulation tasks are selected (an overview is shown in Fig.~\ref{fig:anytask_visual}), covering a diverse set of behaviors including lifting, pushing, stacking, pick-and-place, and drawer opening. Together, these tasks span short-horizon and long-horizon interactions, articulated objects, and contact-rich manipulation scenarios.

Imperfect behavior priors are some scripted policies~\cite{liang2022code} synthesized by an LLM that invokes an existing skill library. These policies provide incomplete state coverage and lack explicit failure recovery behaviors, which constitutes the source of imperfectness. All scripted policies are shown in Fig.~\ref{fig:scripted_policy}. We collect 1000 demonstrations for each task from the scripted policies.

A state-based diffusion policy is trained to generate action chunks with a horizon of 16, conditioned on a structured state vector. The definitions of the state and action representations are summarized in Table~\ref{tab:anytask_state}. For tasks involving fewer than two rigid objects or no articulated objects, the corresponding state dimensions are zero-masked. Task specification is encoded via a learned task embedding, obtained as the output of a lightweight MLP applied to a one-hot task index.
\begin{table*}[htb!]
    \centering
    \begin{tabular}{l|c}
    \toprule
        State &  Dimension \\
        \midrule
        End-effector pose & 16 \\
        robot arm joint positions & 9 \\
        Two object positions & 6 \\ 
        Two object 6D rotations & 12 \\
        Articulated object position & 3 \\
        Articulated object 6D rotation & 6 \\
        Articulated joint position & 1 \\
        Task embedding & 16 \\
        \midrule
        Action &  Dimension \\
        \midrule
        End-effector position & 3 \\
        End-effector 6D rotation & 6 \\
        Gripper action & 1 \\ 
    \bottomrule
    \end{tabular}
    \caption{State condition and action definitions for state-based diffusion policy for the \textsc{AnyTask} benchmark.}
    \label{tab:anytask_state}
\end{table*}

\subsection{AutoMate}
\begin{figure}[hbt!]
    \centering
    \includegraphics[width=0.5\linewidth]{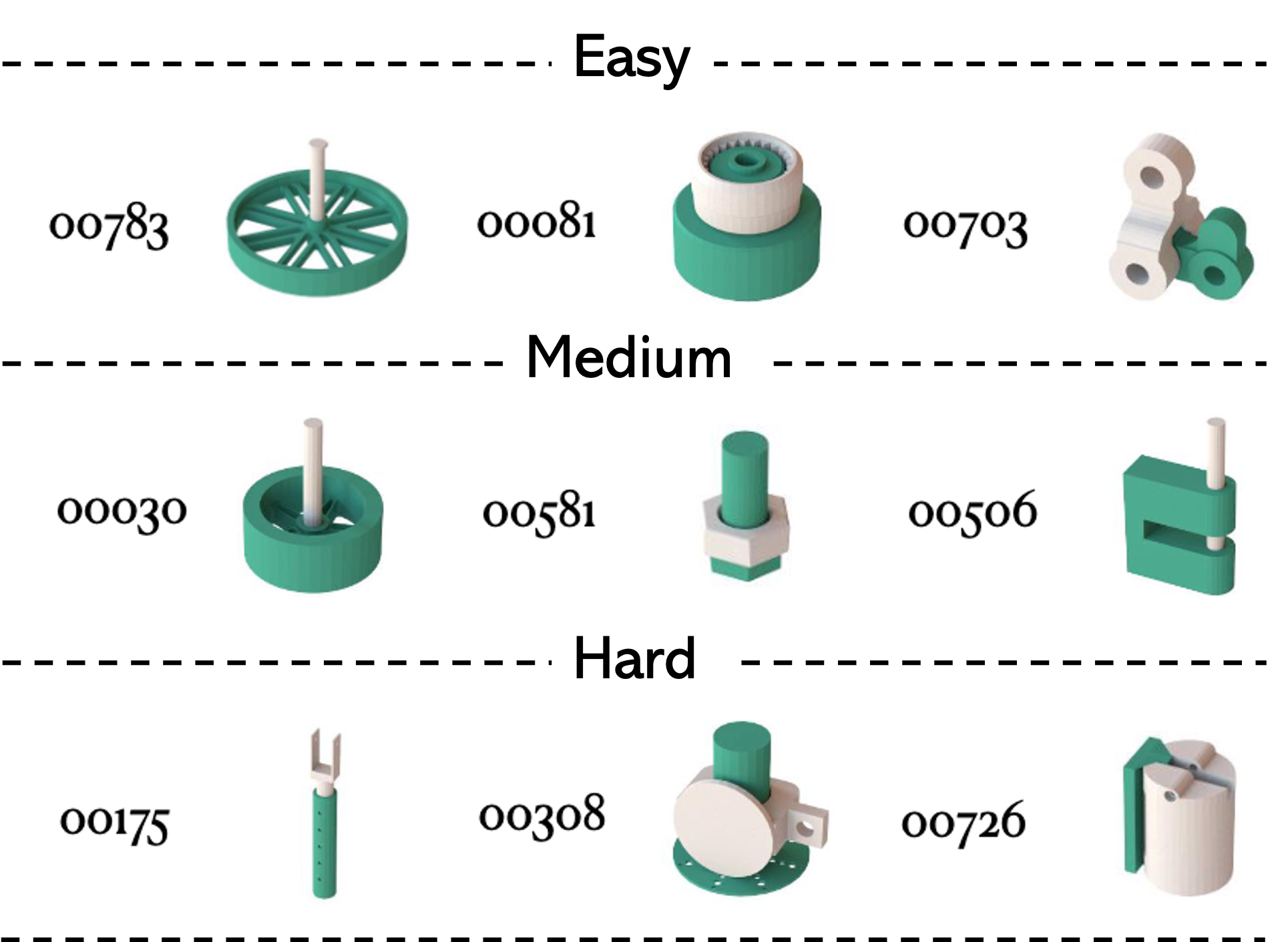}
    \caption{Visualization of the selected nine assets from AutoMate benchmark covering three difficulty levels (easy, medium, and hard).}
    \label{fig:automate_visual}
\end{figure}
AutoMate~\cite{tang2024automate} is a simulation benchmark for robotic dexterous manipulation in industrial assembly, with a particular emphasis on high-precision peg-insertion tasks. The benchmark consists of a collection of insertion scenarios with tight geometric tolerances, demanding accurate pose alignment, fine-grained contact reasoning, and force-sensitive control during execution. For evaluation, nine assets are selected from AutoMate, evenly spanning three difficulty levels: easy (success rate
$>$ 80\%), medium (success rate between 40\% and 80\%), and hard (success rate $<$ 40\%), where difficulty is defined based on the performance of single-task expert policies provided by AutoMate. An overview of the selected assets is shown in Fig.~\ref{fig:automate_visual}.

In this benchmark, imperfect behavior priors are generated from a scripted policy that performs peg disassembly starting from an already assembled configuration. Imperfect demonstrations are obtained by reversing these disassembly trajectories to approximate the assembly process. Because assembly and disassembly dynamics are inherently asymmetric, this inversion induces a dynamics mismatch, resulting in demonstrations with imperfect contact interactions. We collect 500 demonstrations for each task from the scripted policy. The scripted policy for disassembly is shown in Fig.~\ref{fig:automate_scripted} including three skill functions, \texttt{\_disassemble\_plug\_from socket}, \texttt{\_lift\_gripper}, and \texttt{\_randomize\_gripper pose}.

A state-based diffusion policy is trained to generate action chunks with a horizon of 8, conditioned on a structured state vector. The definitions of the state and action representations are summarized in Table~\ref{tab:automate_state}. The 32-dimensional task embedding is computed from a pretrained point-cloud autoencoder similar to AutoMate~\cite{tang2024automate}.

\subsection{Real-World Evaluation}
\label{appendix:real}
Our sim-to-real policies are deployed on a single-arm Franka robot equipped with a Robotiq 2F-85 gripper. Perception is provided by three calibrated Intel RealSense D435i cameras, which yield a merged point-cloud and RGB observations. The RGB observations additionally includes a RealSense D405 wrist mounted camera. All policy inference is performed on a local workstation with an AMD Ryzen Threadripper PRO 5955WX CPU and an NVIDIA RTX 6000 Ada GPU.

\begin{table*}[htb!]
    \centering
    \begin{tabular}{l|c}
    \toprule
        State &  Dimension \\
        \midrule
        End-effector pose & 16 \\
        Held positions & 3 \\ 
        Held 6D rotation & 6 \\
        Task embedding & 32 \\
        \midrule
        Action &  Dimension \\
        \midrule
        End-effector position & 3 \\
        End-effector 6D rotation & 6 \\
    \bottomrule
    \end{tabular}
    \caption{State condition and action definitions for state-based diffusion policy for the AutoMate benchmark.}
    \label{tab:automate_state}
\end{table*}

\small

\begin{figure*}[t]
\centering
\begin{tcolorbox}[
  colback=gray!10,
  colframe=gray!60,
  title=\textbf{Scripted policy - \textit{Lift Banana}},
  fonttitle=\small\sffamily,
  fontupper=\ttfamily\small,
  boxrule=0.5pt,
  arc=2pt
]
\footnotesize
def scripted\_policy(env):\\ \hspace*{1em}    import torch\\ \hspace*{1em}    \\ \hspace*{1em}    \# Object id for banana\\ \hspace*{1em}    banana\_id = 1\\ \hspace*{1em}\\ \hspace*{1em}    \# Step 1: Pick up the banana. This skill will grasp the banana and lift it with an internally defined distance.\\ \hspace*{1em}    pick(env, banana\_id)\\ \hspace*{1em}\\ \hspace*{1em}    \# Step 2: Compute the target position for lifting.\\ \hspace*{1em}    \# We want to ensure the banana is lifted upward by at least 20 cm relative to its initial position.\\ \hspace*{1em}    \# Get the initial position of the banana (batched tensor of shape (N, 3)).\\ \hspace*{1em}    initial\_banana\_pos = get\_object\_initial\_position(env, banana\_id)\\ \hspace*{1em}    \\ \hspace*{1em}    \# Define a lift offset of 0.25 m (25 cm) to ensure it meets the 20 cm requirement.\\ \hspace*{1em}    lift\_offset = torch.tensor([0.0, 0.0, 0.25], device=initial\_banana\_pos.device).reshape(1, 3)\\ \hspace*{1em}    \\ \hspace*{1em}    \# Broadcast the lift offset and add to the initial banana position\\ \hspace*{1em}    target\_pos = initial\_banana\_pos + lift\_offset\\ \hspace*{1em}\\ \hspace*{1em}    \# Step 3: Move the robot's end effector to the target position while keeping the gripper closed.\\ \hspace*{1em}    move\_to(env, target\_pos, target\_orientation=None, gripper\_open=False)\\ \hspace*{1em}\\ \hspace*{1em}    \# Step 4: Ensure the gripper remains closed to keep a firm grasp.\\ \hspace*{1em}    close\_gripper(env)\\ \hspace*{1em}
\end{tcolorbox}
\caption{Scripted policy for \textit{Lift Banana}}
\label{fig:scripted_policy}
\end{figure*}

\begin{figure*}[t]
\centering
\begin{tcolorbox}[
  colback=gray!10,
  colframe=gray!60,
  title=\textbf{Scripted policy - \textit{Lift Brick}},
  fonttitle=\small\sffamily,
  fontupper=\ttfamily\small,
  boxrule=0.5pt,
  arc=2pt
]
\footnotesize
def scripted\_policy(env):\\ \hspace*{1em}    import torch\\ \hspace*{1em}    \\ \hspace*{1em}    \# Object id for soft\_brick\\ \hspace*{1em}    soft\_brick\_id = 1\\ \hspace*{1em}\\ \hspace*{1em}    \# Step 1: Pick up the soft\_brick. This skill will grasp the soft\_brick and lift it with an internally defined distance.\\ \hspace*{1em}    pick(env, soft\_brick\_id)\\ \hspace*{1em}\\ \hspace*{1em}    \# Step 2: Compute the target position for lifting.\\ \hspace*{1em}    \# We want to ensure the soft\_brick is lifted upward by at least 20 cm relative to its initial position.\\ \hspace*{1em}    \# Get the initial position of the soft\_brick (batched tensor of shape (N, 3)).\\ \hspace*{1em}    initial\_soft\_brick\_pos = get\_object\_initial\_position(env, soft\_brick\_id)\\ \hspace*{1em}    \\ \hspace*{1em}    \# Define a lift offset of 0.25 m (25 cm) to ensure it meets the 20 cm requirement.\\ \hspace*{1em}    lift\_offset = torch.tensor([0.0, 0.0, 0.25], device=initial\_soft\_brick\_pos.device).reshape(1, 3)\\ \hspace*{1em}    \\ \hspace*{1em}    \# Broadcast the lift offset and add to the initial soft\_brick position\\ \hspace*{1em}    target\_pos = initial\_soft\_brick\_pos + lift\_offset\\ \hspace*{1em}\\ \hspace*{1em}    \# Step 3: Move the robot\u0027s end effector to the target position while keeping the gripper closed.\\ \hspace*{1em}    move\_to(env, target\_pos, target\_orientation=None, gripper\_open=False)\\ \hspace*{1em}\\ \hspace*{1em}    \# Step 4: Ensure the gripper remains closed to keep a firm grasp.\\ \hspace*{1em}    close\_gripper(env)\\ \hspace*{1em}
\end{tcolorbox}
\caption{Scripted policy for \textit{Lift Brick}}
\label{fig:scripted_policy}
\end{figure*}

\begin{figure*}[t]
\centering
\begin{tcolorbox}[
  colback=gray!10,
  colframe=gray!60,
  title=\textbf{Scripted policy - \textit{Lift Peach}},
  fonttitle=\small\sffamily,
  fontupper=\ttfamily\small,
  boxrule=0.5pt,
  arc=2pt
]
\footnotesize
def scripted\_policy(env):\\ \hspace*{1em}    import torch\\ \hspace*{1em}    \\ \hspace*{1em}    \# Object id for peach\\ \hspace*{1em}    peach\_id = 1\\ \hspace*{1em}\\ \hspace*{1em}    \# Step 1: Pick up the peach. This skill will grasp the peach and lift it with an internally defined distance.\\ \hspace*{1em}    pick(env, peach\_id)\\ \hspace*{1em}\\ \hspace*{1em}    \# Step 2: Compute the target position for lifting.\\ \hspace*{1em}    \# We want to ensure the peach is lifted upward by at least 20 cm relative to its initial position.\\ \hspace*{1em}    \# Get the initial position of the peach (batched tensor of shape (N, 3)).\\ \hspace*{1em}    initial\_peach\_pos = get\_object\_initial\_position(env, peach\_id)\\ \hspace*{1em}    \\ \hspace*{1em}    \# Define a lift offset of 0.25 m (25 cm) to ensure it meets the 20 cm requirement.\\ \hspace*{1em}    lift\_offset = torch.tensor([0.0, 0.0, 0.25], device=initial\_peach\_pos.device).reshape(1, 3)\\ \hspace*{1em}    \\ \hspace*{1em}    \# Broadcast the lift offset and add to the initial peach position\\ \hspace*{1em}    target\_pos = initial\_peach\_pos + lift\_offset\\ \hspace*{1em}\\ \hspace*{1em}    \# Step 3: Move the robot\u0027s end effector to the target position while keeping the gripper closed.\\ \hspace*{1em}    move\_to(env, target\_pos, target\_orientation=None, gripper\_open=False)\\ \hspace*{1em}\\ \hspace*{1em}    \# Step 4: Ensure the gripper remains closed to keep a firm grasp.\\ \hspace*{1em}    close\_gripper(env)
\end{tcolorbox}
\caption{Scripted policy for \textit{Lift Peach}}
\label{fig:scripted_policy}
\end{figure*}

\begin{figure*}[t]
\centering
\begin{tcolorbox}[
  colback=gray!10,
  colframe=gray!60,
  title=\textbf{Scripted policy - \textit{Open Drawer}},
  fonttitle=\small\sffamily,
  fontupper=\ttfamily\small,
  boxrule=0.5pt,
  arc=2pt
]
\footnotesize
def scripted\_policy(env):\\ \hspace*{1em}   obj\_id = 1 
\\ \hspace*{1em}    open\_drawer(env, obj\_id)
\end{tcolorbox}
\caption{Scripted policy for \textit{Open Drawer}}
\label{fig:scripted_policy}
\end{figure*}

\begin{figure*}[t]
\centering
\begin{tcolorbox}[
  colback=gray!10,
  colframe=gray!60,
  title=\textbf{Scripted policy - \textit{Push Pear to Center}},
  fonttitle=\small\sffamily,
  fontupper=\ttfamily\small,
  boxrule=0.5pt,
  arc=2pt
]
\footnotesize
def scripted\_policy(env):
\\ \hspace*{1em} import torch
\\ \hspace*{1em}    \# Push the pear (object\_id=1) to target xy (0.4, -0.5) using iterative, state-based pushes\\ \hspace*{1em}    pear\_id = 1\\ \hspace*{1em}    target\_xy = make\_a\_tensor(env, [0.4, -0.44])\\ \hspace*{1em}    push\_or\_pull\_object\_to\_xy(env, pear\_id, target\_xy)
\end{tcolorbox}
\caption{Scripted policy for \textit{Push Pear to Center}}
\label{fig:scripted_policy}
\end{figure*}

\begin{figure*}[t]
\centering
\begin{tcolorbox}[
  colback=gray!10,
  colframe=gray!60,
  title=\textbf{Scripted policy - \textit{Stack Banana on Can}},
  fonttitle=\small\sffamily,
  fontupper=\ttfamily\small,
  boxrule=0.5pt,
  arc=2pt
]
\footnotesize
def scripted\_policy(env):\\ \hspace*{1em}    import torch\\ \hspace*{1em}    banana\_id = 2\\ \hspace*{1em}    can\_id = 1\\ \hspace*{1em}\\ \hspace*{1em}    \# 1) Pick the banana\\ \hspace*{1em}    pick(env, banana\_id)\\ \hspace*{1em}\\ \hspace*{1em}    \# 2) Lift to 25 cm above initial height\\ \hspace*{1em}    init\_banana = get\_object\_initial\_position(env, banana\_id)  \# (N,3)\\ \hspace*{1em}    lift\_offset = torch.tensor([0.0, 0.0, 0.18], device=init\_banana.device).reshape(1, 3)\\ \hspace*{1em}    lifted\_pos = init\_banana + lift\_offset\\ \hspace*{1em}    move\_to(env, lifted\_pos, target\_orientation=None, gripper\_open=False)\\ \hspace*{1em}    can\_position = get\_object\_position(env, can\_id)\\ \hspace*{1em}    target\_position = can\_position.clone()\\ \hspace*{1em}    place(env, target\_position)
\end{tcolorbox}
\caption{Scripted policy for \textit{Stack Banana on Can}}
\label{fig:scripted_policy}
\end{figure*}

\begin{figure*}[t]
\centering
\begin{tcolorbox}[
  colback=gray!10,
  colframe=gray!60,
  title=\textbf{Scripted policy - \textit{Put Object in Closed Drawer}},
  fonttitle=\small\sffamily,
  fontupper=\ttfamily\small,
  boxrule=0.5pt,
  arc=2pt
]
\footnotesize
def scripted\_policy(env):\\ \hspace*{1em}    import torch\\ \hspace*{1em}    obj\_id = 1  \# drawer\\ \hspace*{1em}    strawberry\_id = 2\\ \hspace*{1em}\\ \hspace*{1em}    \# 1) Open the upper drawer\\ \hspace*{1em}    open\_drawer(env, obj\_id, reset\_after\_open=True)\\ \hspace*{1em}\\ \hspace*{1em}    \# 2) Pick the strawberry\\ \hspace*{1em}    pick(env, strawberry\_id)\\ \hspace*{1em}\\ \hspace*{1em}    \# 3) Lift strawberry ~23 cm above its initial position\\ \hspace*{1em}    init\_straw = get\_object\_initial\_position(env, strawberry\_id)  \# (N,3)\\ \hspace*{1em}    lift\_offset = torch.tensor([0.0, 0.0, 0.23], device=init\_straw.device).reshape(1, 3)\\ \hspace*{1em}    lifted\_pos = init\_straw + lift\_offset\\ \hspace*{1em}    move\_to(env, lifted\_pos, target\_orientation=None, gripper\_open=False)\\ \hspace*{1em}\\ \hspace*{1em}    \# 4) Place location: use upper drawer's (x, y) and fixed z = 0.23\\ \hspace*{1em}    drawer\_pos = get\_articulated\_obj\_part\_pos(env, obj\_id, 'upper\_drawer', \_use\_bbox\_center=True)  \# (N,3)\\ \hspace*{1em}    z\_const = torch.full((drawer\_pos.shape[0], 1), 0.23, device=drawer\_pos.device)\\ \hspace*{1em}    place\_target = torch.cat([drawer\_pos[:, :2], z\_const], dim=1)  \# (N,3)\\ \hspace*{1em}\\ \hspace*{1em}    \# Move to place location and open gripper directly (no pre-place)\\ \hspace*{1em}    move\_to(env, place\_target, target\_orientation=None, gripper\_open=False)\\ \hspace*{1em}    open\_gripper(env)
\end{tcolorbox}
\caption{Scripted policy for \textit{Put Object in Closed Drawer}}
\label{fig:scripted_policy}
\end{figure*}

\begin{figure*}[t]
\centering
\begin{tcolorbox}[
  colback=gray!10,
  colframe=gray!60,
  title=\textbf{Scripted policy - \textit{Place Strawberry in Bowl}},
  fonttitle=\small\sffamily,
  fontupper=\ttfamily\small,
  boxrule=0.5pt,
  arc=2pt
]
\footnotesize
def scripted\_policy(env):\\ \hspace*{1em}     import torch\\ \hspace*{1em}     strawberry\_id = 2\\ \hspace*{1em}     bowl\_id = 1\\ \hspace*{1em} \\ \hspace*{1em}     \# 1) Pick the strawberry\\ \hspace*{1em}     pick(env, strawberry\_id)\\ \hspace*{1em} \\ \hspace*{1em}     \# 2) Lift to 25 cm above initial height\\ \hspace*{1em}     init\_strawberry = get\_object\_initial\_position(env, strawberry\_id)  \# (N,3)\\ \hspace*{1em}     lift\_offset = torch.tensor([0.0, 0.0, 0.18], device=init\_strawberry.device).reshape(1, 3)\\ \hspace*{1em}     lifted\_pos = init\_strawberry + lift\_offset\\ \hspace*{1em}     move\_to(env, lifted\_pos, target\_orientation=None, gripper\_open=False)\\ \hspace*{1em}     bowl\_position = get\_object\_position(env, bowl\_id)\\ \hspace*{1em}     target\_position = bowl\_position.clone()\\ \hspace*{1em}     place(env, target\_position)
\end{tcolorbox}
\caption{Scripted policy for \textit{Put Object in Closed Drawer}}
\label{fig:scripted_policy}
\end{figure*}

\begin{figure*}[t]
\centering
\begin{tcolorbox}[
  colback=gray!10,
  colframe=gray!60,
  title=\textbf{Scripted policy - \_disassemble\_plug\_from\_socket},
  fonttitle=\small\sffamily,
  fontupper=\ttfamily\small,
  boxrule=0.5pt,
  arc=2pt
]
\footnotesize
def \_disassemble\_plug\_from\_socket(self):
\\ \hspace*{1em}"""Lift plug from socket till disassembly and then randomize end-effector pose."""
\\
\\ \hspace*{1em}if\_intersect = np.ones(self.num\_envs, dtype=np.float32)
\\
\\ \hspace*{1em}env\_ids = np.argwhere(if\_intersect == 1).reshape(-1)
\\ \hspace*{1em}lift\_distance = self.disassembly\_dists * 3.0
\\ \hspace*{1em}self.\_lift\_gripper(lift\_distance, self.cfg\_task.disassemble\_sim\_steps, env\_ids)
\\
\\ \hspace*{1em}self.step\_sim\_no\_action()
\\
\\ \hspace*{1em}if\_intersect = (self.held\_pos[:, 2] < self.fixed\_pos[:, 2] + self.disassembly\_dists).cpu().numpy()
\\ \hspace*{1em}env\_ids = np.argwhere(if\_intersect == 0).reshape(-1)
\\ \hspace*{1em}self.\_randomize\_gripper\_pose(self.cfg\_task.move\_gripper\_sim\_steps, env\_ids)
\end{tcolorbox}

\begin{tcolorbox}[
  colback=gray!10,
  colframe=gray!60,
  title=\textbf{Scripted policy - \_lift\_gripper},
  fonttitle=\small\sffamily,
  fontupper=\ttfamily\small,
  boxrule=0.5pt,
  arc=2pt
]
\footnotesize
def \_lift\_gripper(self, lift\_distance, sim\_steps, env\_ids=None):
\\ \hspace*{1em}"""Lift gripper by specified distance. Called outside RL loop (i.e., after last step of  
\\ \hspace*{1em}episode)."""
\\
\\ \hspace*{1em}ctrl\_tgt\_pos = torch.empty\_like(self.fingertip\_midpoint\_pos).\\copy\_(self.fingertip\_midpoint\_pos)
\\ \hspace*{1em}ctrl\_tgt\_quat = torch.empty\_like(self.fingertip\_midpoint\_quat).\\copy\_(self.fingertip\_midpoint\_quat)
\\ \hspace*{1em}ctrl\_tgt\_pos[:, 2] += lift\_distance
\\ \hspace*{1em}lifted = self.fixed\_pos[:, 2] > self.disassembly\_dists * 0.95
\\ \hspace*{1em}ctrl\_tgt\_pos[lifted, :2] += torch.randn((self.num\_envs, 2), dtype=torch.float32, device=self.device) * 0.02
\\ \hspace*{1em}if len(env\_ids) == 0:
\\ \hspace*{1em}\hspace*{1em}env\_ids = np.array(range(self.num\_envs)).reshape(-1)
\\
\\ \hspace*{1em}self.\_move\_gripper\_to\_eef\_pose(env\_ids, ctrl\_tgt\_pos, ctrl\_tgt\_quat, sim\_steps, if\_log=True)
\end{tcolorbox}

\caption{Scripted policy for AutoMate disassembly tasks.}
\label{fig:automate_scripted}
\end{figure*}

\begin{figure*}[t]
\centering
\begin{tcolorbox}[
  colback=gray!10,
  colframe=gray!60,
  title=\textbf{Scripted policy - \_randomize\_gripper\_pose},
  fonttitle=\small\sffamily,
  fontupper=\ttfamily\small,
  boxrule=0.5pt,
  arc=2pt
]
\footnotesize
def \_randomize\_gripper\_pose(self, sim\_steps, env\_ids):
\\ \hspace*{1em}"""Move gripper to random pose."""
\\
\\ \hspace*{1em}ctrl\_tgt\_pos = torch.empty\_like(self.gripper\_goal\_pos).copy\_( self.gripper\_goal\_pos )
\\ \hspace*{1em}ctrl\_tgt\_pos[:, 2] += self.cfg\_task.gripper\_rand\_z\_offset
\\
\\ \hspace*{1em}\# ctrl\_tgt\_pos = torch.empty\_like(self.fingertip\_midpoint\_pos).copy \_( self.fingertip\_midpoint\_pos )
\\
\\ \hspace*{1em}fingertip\_centered\_pos\_noise = 2 * (
\\ \hspace*{1em}\hspace*{1em}torch.rand((self.num\_envs, 3), dtype=torch.float32, device=self.device) - 0.5
\\ \hspace*{1em})  \# [-1, 1]
\\ \hspace*{1em}pos\_noise = torch.tensor([
\\ \hspace*{1em}\hspace*{1em}0.03, 0.03, 0.02 + self.disassembly\_dists[0]
\\ \hspace*{1em}], device=self.device)
\\ \hspace*{1em}fingertip\_centered\_pos\_noise = fingertip\_centered\_pos\_noise @ torch.diag(pos\_noise)
\\ \hspace*{1em}fingertip\_centered\_pos\_noise[:, 2] -= (0.02 + self.disassembly\_dists[0]) / 2.0
\\ \hspace*{1em}ctrl\_tgt\_pos += fingertip\_centered\_pos\_noise
\\
\\ \hspace*{1em}\# Set target rot
\\ \hspace*{1em}ctrl\_target\_fingertip\_centered\_euler = (
\\ \hspace*{1em}\hspace*{1em} torch.tensor(self.cfg\_task.fingertip\_centered\_rot\_initial, device=self.device)
\\ \hspace*{1em}\hspace*{1em}.unsqueeze(0)
\\ \hspace*{1em}\hspace*{1em}.repeat(self.num\_envs, 1)
\\ \hspace*{1em})
\\
\\ \hspace*{1em}fingertip\_centered\_rot\_noise = 2 * (
\\ \hspace*{1em}\hspace*{1em}torch.rand((self.num\_envs, 3), dtype=torch.float32, device=self.device) - 0.5
\\ \hspace*{1em})  \# [-1, 1]
\\ \hspace*{1em}fingertip\_centered\_rot\_noise = fingertip\_centered\_rot\_noise @ torch.diag(
\\ \hspace*{1em}\hspace*{1em} torch.tensor(self.cfg\_task.gripper\_rand\_rot\_noise, device=self.device)
\\ \hspace*{1em})
\\ \hspace*{1em}ctrl\_target\_fingertip\_centered\_euler += fingertip\_centered\_rot\_noise
\\ \hspace*{1em}ctrl\_tgt\_quat = torch\_utils.quat\_from\_euler\_xyz(
\\ \hspace*{1em}\hspace*{1em} ctrl\_target\_fingertip\_centered\_euler[:, 0],
\\ \hspace*{1em}\hspace*{1em} ctrl\_target\_fingertip\_centered\_euler[:, 1],
\\ \hspace*{1em}\hspace*{1em} ctrl\_target\_fingertip\_centered\_euler[:, 2],
\\ \hspace*{1em})
\\
\\ \hspace*{1em}self.\_move\_gripper\_to\_eef\_pose(env\_ids, ctrl\_tgt\_pos, ctrl\_tgt\_quat, sim\_steps, if\_log=True)
\end{tcolorbox}
\caption{Scripted policy for AutoMate disassembly tasks.}
\label{fig:automate_scripted}
\end{figure*}

\end{document}